\pdfoutput=1
\documentclass[11pt]{article}

\usepackage[final]{acl}
\usepackage{booktabs}
\usepackage{multirow} % Required for multirow feature
\usepackage{times}
\usepackage{graphicx}
\usepackage{subcaption}
\usepackage{latexsym}
\usepackage[inline]{enumitem}

\usepackage[T1]{fontenc}
\usepackage[utf8]{inputenc}

\usepackage{microtype}

\usepackage{inconsolata}

\usepackage{graphicx}
\usepackage{xspace}
\usepackage{amsmath} 
\usepackage[nameinlink]{cleveref}

\newcommand{\dataset}{CS4\xspace}
\newcommand{\datasetnospace}{CS4}

\newcommand{\expsec}[1]{\vspace{+0.2cm} \noindent \textcolor{violet}{\underline{\textbf{#1}}}} 

\newcommand{\att}[1]{\noindent \textcolor{violet}{\underline{\textbf{#1}}}} 

\title{\dataset: Measuring the Creativity of Large Language Models Automatically \\ by Controlling the Number of Story-Writing Constraints}

\author{  
Anirudh Atmakuru\thanks{\, indicates equal contribution.} \, \, \,
Jatin Nainani\footnotemark[1] \, \, \,
Rohith Siddhartha Reddy Bheemreddy\footnotemark[1] 
\\ 
{\bf Anirudh Lakkaraju}\footnotemark[1] \, \, \,
{\bf Zonghai Yao} \, \, \,
{\bf Hamed Zamani} \, \, \,
{\bf Haw-Shiuan Chang}\footnotemark[1]\thanks{\, The work was mostly done at Amazon.} \\ 
  University of Massachusetts, Amherst\\
  140 Governors Dr., Amherst, MA, USA 01003\\
  \texttt{ \{aatmakuru, jnainani, rbheemreddy, alakkaraju, zonghaiyao\}@umass.edu } \\ \texttt{\{zamani,hschang\}@cs.umass.edu} 
  }

\begin{document}
\maketitle
\begin{abstract}
Evaluating the creativity of large language models (LLMs) in story writing is difficult because LLM-generated stories could seemingly look creative but be very similar to some existing stories in their huge and proprietary training corpus. To overcome this challenge, we introduce a novel benchmark dataset with varying levels of prompt specificity: \dataset (\textbf{C}omparing the \textbf{S}kill of \textbf{C}reating \textbf{S}tories by \textbf{C}ontrolling the \textbf{S}ynthesized \textbf{C}onstraint \textbf{S}pecificity). By increasing the number of requirements/constraints in the prompt, we can increase the prompt specificity and hinder LLMs from retelling high-quality narratives in their training data. Consequently, \dataset empowers us to indirectly measure the LLMs' creativity without human annotations. 

Our experiments on LLaMA, Gemma, and Mistral not only highlight the creativity challenges LLMs face when dealing with highly specific prompts but also reveal that different LLMs perform very differently under different numbers of constraints and achieve different balances between the model's instruction-following ability and narrative coherence.
Additionally, our experiments on OLMo suggest that Learning from Human Feedback (LHF) can help LLMs select better stories from their training data but has limited influence in boosting LLMs' ability to produce creative stories that are unseen in the training corpora. The benchmark is released at \url{https://github.com/anirudhlakkaraju/cs4_benchmark}.

\end{abstract}

\noindent “Creativity is seeing what others see and thinking what no one else ever thought.” ― Albert Einstein

\begin{figure}[t!]
  \centering
  \includegraphics[width=0.9\columnwidth]{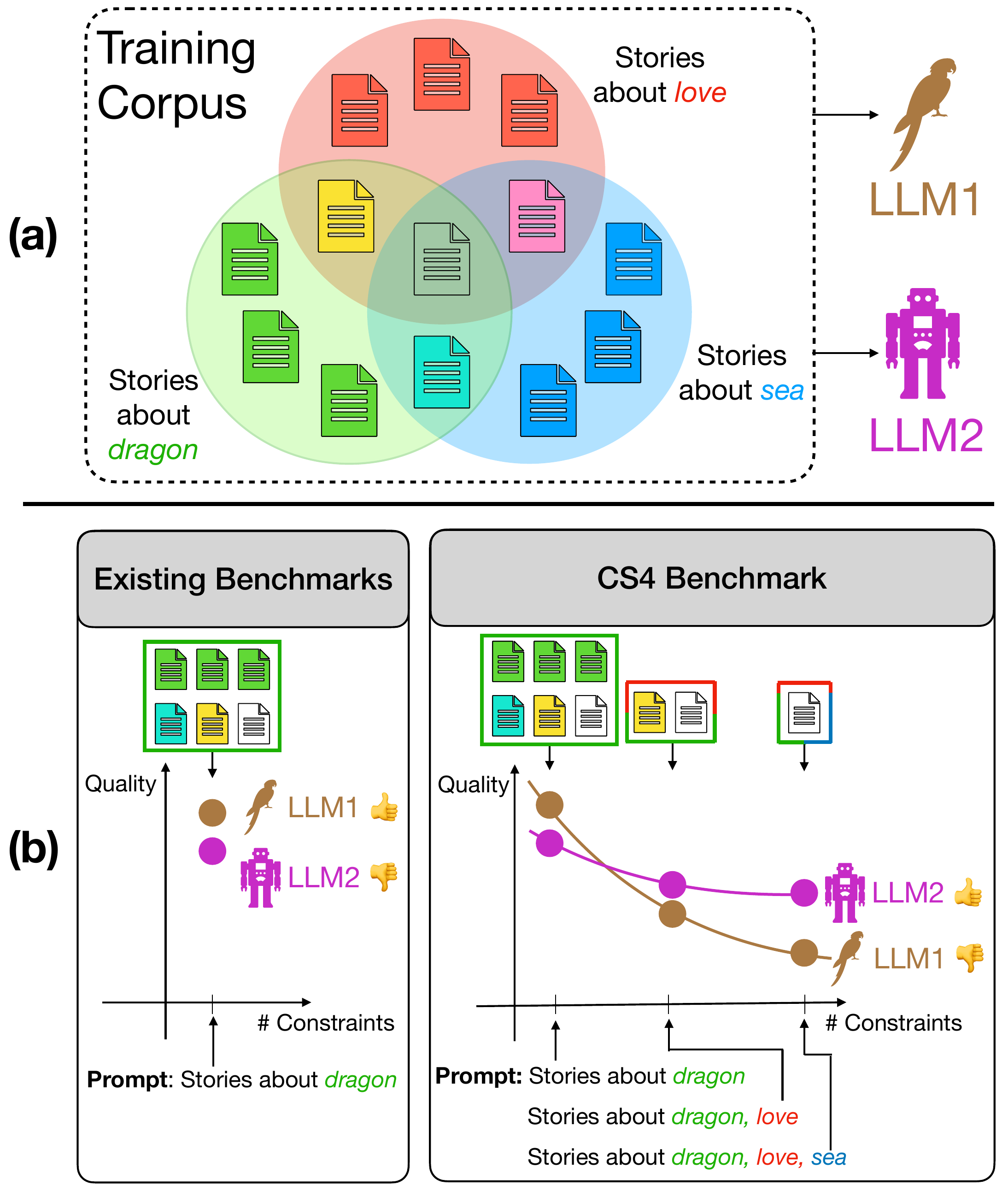}
  \caption{Comparison between \dataset and existing benchmarks. (a) Depiction of training corpora subsets for different narrative themes, illustrating the decreasing availability of training examples for LLMs as prompt specificity increases. (b) In response to general instructions, LLM1 tends to copy the relevant high-quality stories from its training corpus to achieve a good score in existing story-writing benchmarks. \dataset measures LLMs’ creativity by comparing LLMs’ performance drops for more specific instructions. Given more constraints, LLM2 could leverage very limited training data to output higher-quality stories than LLM1, so LLM2 is more creative.}
  \label{fig:Venn_diag}
\end{figure}

%(a) A small subset of LLMs’ training corpus that is relevant to the three story topics.

\section{Introduction}

Large language models (LLMs) can generate stories that surpass the quality of human-written ones, which intensifies debates among researchers and professional writers about whether human writers may be replaced by LLMs anytime soon \citep{gomez2023confederacy, zhao2023more, marco2024pron}. One common argument on the human side is that LLMs still lack creativity~\citep{boussioux2023crowdless,chakrabarty2024art,gomez2023confederacy}, which is defined as the ability to produce original, unseen, and high-quality stories in this study. 

%\citet{ chia2024instructeval, guan2021openmeva, he2024can, jiang2023followbench, qin2024infobench, yao2023collie, yuan2022wordcraft, zhou2023instruction, li2024cif, zhang2024cfbench}, 

Current evaluation methodologies are often insufficient for measuring creativity in story writing. In existing creative-writing or instruction-following benchmarks such as FollowBench~\citep{jiang2023followbench}, IFEval~\citep{zhou2023instruction}, Collie~\citep{yao2023collie}, CIF-Bench~\citep{li2024cif}, InstructEval~\citep{chia2024instructeval}, CFBench~\citep{zhang2024cfbench}, and InfoBench~\citep{qin2024infobench}, instructions are typically broad and impose limited constraints, which allow LLMs to perform well by slightly modifying (or even copying) relevant and high-quality narratives from their training datasets (\citealp{chang2023speak}; \citealp{hartmann2023sok}; \citealp{min2023silo}) without actually ``understanding'' the output story~\citep{west2023generative}. However, it is challenging to directly check if the generated stories mostly come from the stories in the model's training corpora~\citep{chen2024copybench}. This is because 1) LLMs could (or might often) output text that is semantically similar to but lexically different from a training text piece \citep{ippolito2023preventing}, 2) the training corpora are usually huge, and searching these for identifying LLM-generated text could be expensive,
%storing them and retrieving text from it could be expensive
and 3) the training data used for training many popular LLMs are not publicly available.  

To assess the creativity of LLMs in terms of generating original stories, we propose a novel evaluation benchmark, \datasetnospace, 
%(\textbf{C}omparing the \textbf{S}kill of \textbf{C}reating \textbf{S}tories by \textbf{C}ontrolling the \textbf{S}ynthesized \textbf{C}onstraint \textbf{S}pecificity)
which provides instructions with up to $39$ constraints. As we have more constraints in the prompt, the instruction becomes more specific and limits the LLM's ability to copy text from its training data. For example, as shown in \Cref{fig:Venn_diag}, given a prompt like \textit{``Give me a story about a dragon''}, an LLM may reproduce the finest story from its training data that mentions the word \textit{``dragon''}, which could naturally be superior to an average story crafted by a human. However, this might not be the case as the specificity of the prompt increases since the LLM would have fewer training examples to leverage, for example, \textit{``Give me a story involving a dragon, love, and sea''}. 

Adding more constraints to a prompt can not only push LLMs to be more creative but also help us assess how effectively they handle the intricate challenges human writers face \citep{he2024complex,wen2024benchmarking,pham2024suri}. This makes our \dataset benchmark both scientifically valuable and practically relevant for professional writers, particularly in the book and film industries. These writers are tasked with crafting intricate, high-quality narratives that meet various constraints from the plots from the existing chapters~\citep{ippolito2022creative}, from the rules of their imaginary world, and by the different needs of editors, the marketing team, and the readers. Understanding the extent to which LLMs can replicate this process will not only help us deep dive into their behavior but could also be essential for determining their applicability in these creative domains.

%ToDo: start to introduce our methods

Similar to \citet{jiang2023followbench, yao2023collie, zhou2023instruction}, we synthesize our prompts by asking GPT-4~\citep{achiam2023gpt} to generate up to $39$ constraints for the story-writing instructions from users. 
%, which allows us to produce an arbitrary number of constraints. 
The constraints in \dataset are synthesized in two ways: 1) manually writing constraints for some user instructions as few-shot examples, which help GPT-4 generate constraints for a new user instruction, and 2) using GPT-4 to extract the constraints from a human-written story to make sure the constraints are realistic and satisfiable. 

\dataset measures common metrics such as instruction-following ratio, coherence, and diversity. However, unlike the existing benchmarks that provide a single score for each metric, \dataset uses multiple sets of prompts with different numbers of constraints to compare LLMs' performances across a range of prompt specificities (\Cref{fig:Venn_diag}). A more specific prompt (i.e., one with more constraints) is usually more difficult to satisfy, but a more creative LLM should have a smaller performance degradation.

In our experiments, we first found that all the explored LLMs output significantly worse stories as the prompt specificity increases. This indicates that more constraints indeed pose more creativity challenges on LLMs for generating novel responses ~\citep{chakrabarty2024art,lu2024benchmarking}. Second, given more constraints, we found that the stories from some LLMs deteriorate faster than those from others. For example, increasing the number of constraints from $7$ to $39$ makes the constraint satisfaction probability drop by approximately $35\%$ for LLaMA-2 7B \citep{touvron2023llama} but by only approximately $20\%$ for Gemma-7B \citep{team2024gemma}. Third, not all metrics degrade at the same speed. For example, when we increase the number of constraints from $7$ to $23$, LLaMA-2 outputs the stories that are similarly coherent but satisfy $25\%$ less constraints. Finally, by comparing OLMo Instruct and OLMo SFT~\citep{groeneveld2024olmo}, we found that the performance enhancement due to Direct Preference Optimization (DPO) \citep{rafailov2024direct} or more generally, Learning from Human Feedback (LHF) \citep{ouyang2022training} is much smaller given more constraints. The results support the hypothesis that LHF achieves better performance but worse diversity by encouraging the LLMs to select good stories from the training data~\citep{kirk2024understanding, le2024exploring, xiao2024algorithmic,lake2024distributional}.

\begin{figure*}%[t]
  \includegraphics[width=\linewidth]{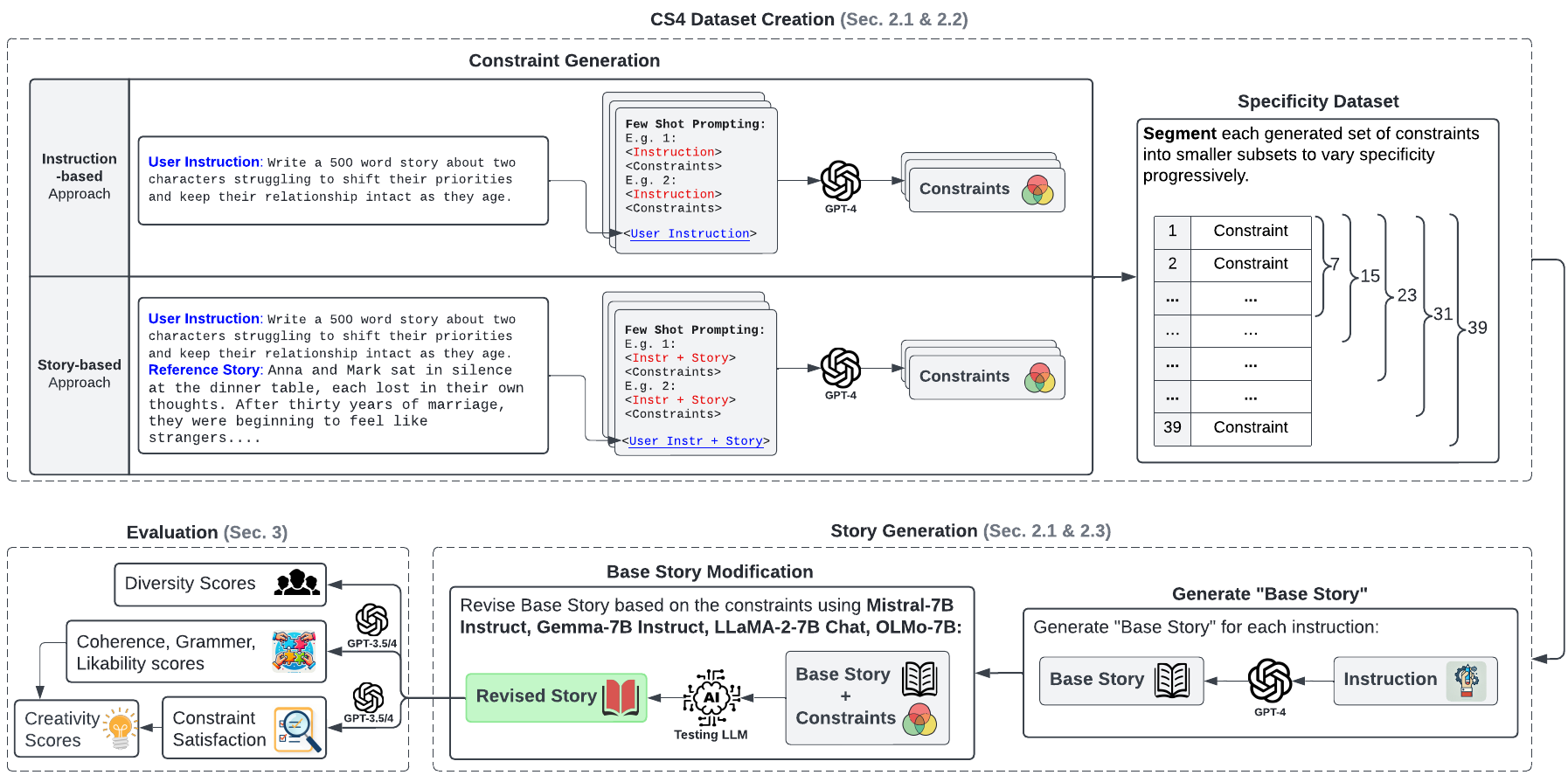}
  \caption{An overview of the evaluation process in \dataset benchmark. First, we use two different few-shot in-context learning approaches to synthesize 39 constraints from every user instruction and conduct sub-sampling to create the prompts with fewer constraints. Next, for each user instruction, the testing LLMs of interest revise a base story, which is generated without seeing the constraints, to satisfy the constraints. The revised stories are evaluated in terms of their constraint satisfaction ratio, quality, and diversity. Finally, we estimate LLMs' creativity by summarizing their coherence scores and instruction satisfaction ratios for different number of constraints.}
  \label{fig:Methodology}
\end{figure*}

\section{Benchmark Construction}
As illustrated in \Cref{fig:Methodology}, the methodology adopted in this research can be broken down into three steps – 1) generating constraints, 2) generating stories, and 3) evaluating the stories based on the constraints. In this section, we first explain the challenges encountered in constructing the benchmark and the solution we propose in each step to overcome these challenges. Then, we describe some important details of the first two steps.

\subsection{Challenges and Solutions}
\label{sec:challenges}
When constructing the \dataset benchmark, we iteratively improve our evaluation process and prompting strategies. Our goal is to build a benchmark that allows us to observe an LLM's response given many constraints making it challenging for the LLM. In each iteration, we run small-scale experiments on ChatGPT (GPT-3.5-Turbo) and manually evaluate their responses. We summarize the challenges faced and our solutions below. The challenges also highlight the differences between \dataset and previous studies.

\expsec{Generating lots of constraints:} 
In the previous instruction-following studies, the number of constraints for each user instruction is usually fixed and small. For example, even the ``hard'' set of InfoBench~\citep{qin2024infobench} only contains $6.3$ constraints on average and each instruction has only one set of constraints. 

In \datasetnospace, we first synthesize many constraints for every user instruction using GPT-4~\citep{achiam2023gpt} to progressively sample the intervals from these constraints later to control the number of constraints of each instruction. We call this method the instruction-based approach because GPT-4 generates several constraints based only on the user instruction.

\expsec{Not generating very difficult constraints:} 
When there are many synthesized constraints, the constraints might contradict each other, making some constraints unsatisfiable while writing a story. In the instruction-based approach, we prompt GPT-4 to make the constraints satisfiable. Besides, we propose a story-based approach, which asks GPT-4 to derive the constraints from the user instruction and a reference story that satisfies the instruction~\citep{li2024self,pham2024suri}, to further address this issue.

In both approaches, our prompts encourage the GPT-4 to generate atomic constraints \citep{min2023factscore}, which means each constraint cannot be broken down further into simpler parts. This alleviates the problem of generating very complex constraints that are hard to follow.

\expsec{Not generating very easy constraints:}
%An instruction following benchmark that is too easy for LLMs might not differentiate LLMs' capability well
An instruction following benchmark that is easy for LLMs may have low discriminative power for LLMs' creative capabilities~\citep{jiang2023followbench,zhang2024cfbench}. For prompts generated using the instruction-based approach, we observe that ChatGPT tends to output fantasy stories. Since anything could happen in a fantasy world, LLMs could easily satisfy the arbitrary number of constraints. To solve this problem, we restrict the user instructions to queries on realistic fictions, which force LLMs to respect common sense and logical rules in the real world.
%impose implicitly common sense and logical constraints.

For the story-based approach, our first version of the prompt makes GPT-4 sequentially modify each sentence in the reference story as a constraint. This would allow a story writer to simply convert the constraints back to sentences, forming a story very similar to the reference story. To address this issue, we prompt GPT-4 to generate constraints that span multiple sentences of the story and encourage the production of constraints that do not follow the order of the story. In \Cref{fig:Type2:OldVNew}, we compare the constraints before and after addressing this issue.

\expsec{Reducing the instability of evaluation:}
It is difficult to find a metric that can objectively compare the quality of two stories from two different genres. For example, incoherence or logical flaws might be acceptable in an adventure story but disastrous in a detective story. Some human or LLM evaluators might also prefer one genre over another. To reduce the bias and variance in the evaluation, we first ask GPT-4 to generate a ``base'' story for the user instruction without seeing the constraints. Then, the 7B-LLMs that are being evaluated are prompted to modify the base story to satisfy the constraints. Because of the similarity of output stories from different testing LLMs, it is easier to compare their qualities.

The extra base story generation step also prevents the tested LLMs from ``cheating''. The LLMs might have seen and memorized the reference story used in the story-based approach and the base story prevents the LLMs from outputting the reference story. Besides, without the base story, LLMs might output a story that contains just the copied input constraints from the instruction-based prompt with some simple transitions between constraints.

\subsection{\dataset Dataset Creation}
\label{sec: Dataset Creation}
As illustrated in \Cref{fig:Methodology} and discussed in the previous section, we use two strategies to synthesize the constraints and enrich the diversity of the \dataset benchmark. For each constraint generation strategy, our benchmark dataset consists of $50$ user instructions.

%Figures~\ref{fig:D2 Constraints} and ~\ref{fig:Type2:OldVNew} show examples of constraints generated using both these strategies.

\expsec{Instruction-based Approach:}
We collect $50$ realistic fiction ideas from Kindlepreneur\footnote{\url{https://kindlepreneur.com/realistic-fiction-story-ideas/}} as user instructions and manually create two few-shot examples of instructions with ten constraints each. These in-context examples guide GPT-4 in generating high-quality constraints, which include more style-related or open-ended elements compared to the story-based approach. Please see \Cref{fig:D2 Constraints} for an example of such constraints.

%find that in-context learning are helpful for GPT-4 to increase the constraint quality.
%Compared to the story-based approach, instruction-based approach could generate more style-related or open-ended constraints and less plot-related constraints.

\expsec{Story-based Approach:}
We collect $50$ user instructions and their corresponding human-written reference stories from the Writing Prompts dataset \citep{fan2018hierarchical} and our manually written few-shot examples encourage GPT-4 to extract more abstract and non-sequential constraints from the reference story. Compared to the instruction-based approach, this method synthesizes more plot-related constraints.

%story-based constraints were written based on the latest stories obtained from .
%(Please see an example that compare the generated constraints before and after solving this problem in ).

\expsec{Constraint Segmentation:}
\label{sec:constraint_num}
For each instruction from the above constraint generation strategies, we segment sets of $7$, $15$, $23$, $31$, and $39$ constraints cumulatively. This segmentation results in $500$ unique prompts that could be used for story generation ($50$ instructions $*$ $2$ constraint generation approaches $*$ $5$ sets of constraints).

\subsection{Story Generation}
We first prompt GPT-4 to write a ``base'' story with less than $500$ words given the user instruction alone. Then, we provide the instruction, the base story, and the constraints to the chosen 7B-LLMs asking them to modify the base story such that the constraints are satisfied in the new story in about $500$ words. This constraint on the story length increases the task difficulty and reduces the costs of the subsequent LLM evaluation.

\section{Evaluation Metrics}
\label{sec:Evaluation Metrics}
In this work, we primarily use four metrics to evaluate the stories generated. These include constraint satisfaction, story coherence, output diversity, and LLM's creativity. In the interest of evaluation cost, we adopt LLM-as-the-Judge~\citep{liu2023g} and use GPT-3.5-Turbo to evaluate constraint satisfaction and story coherence by default. Besides the scores, we also ask GPT-3.5-Turbo to provide justifications to improve and support its evaluation judgement~\citep{chiang2023closer}.

\subsection{Constraint Satisfaction}
\label{sec:constraint-satisfaction}

We use GPT-3.5-Turbo to judge whether the responses from the testing LLMs satisfy each constraint in the prompt~\citep{jiang2023followbench,qin2024infobench,chia2024instructeval,zhang2024cfbench,wen2024benchmarking}. Then, our constraint satisfaction metric is defined as the ratio of the number of constraints satisfied by the story to the total number of constraints in the prompt. You can find our prompt in \Cref{fig: Constraint Satisfaction Eval Prompt}. Our preliminary evaluations show that the judgment of GPT-3.5-Turbo is highly correlated with human judgments in \datasetnospace.

\subsection{Story Coherence}
\label{sec:Story Coherence and the Linear Evaluation Algorithm}

Adding more constraints tends to make the stories incoherent. Further, coherence is a more objective metric than likability, so we use coherence as our main metric to evaluate the generated story. In \Cref{sec:grammar_likability}, we also report the results of grammar and likeability. We evaluate coherence using GPT-3.5-Turbo, which has been shown to match the performance of human evaluators \citep{gilardi2023chatgpt}.

We found that GPT-3.5-Turbo tends to be too generous (i.e., often assigning the highest score) when evaluating the stories generated by LLMs in \dataset \citep{gmyrek2024technological}. To solve the issue, we choose a generated story with the middle number of constraints ($23$ in our experiments) as a baseline for every user instruction, and compare all stories for that user instruction with the baseline. To avoid the positional bias, we randomly swap the order of each generated story and the baseline in the comparison~\citep{zeng2023evaluating}. This evaluation method reduces the times of running GPT-3.5-Turbo from the $N^2$ to $N-1$ compared to the exhaustive pairwise comparison~\citep{liusie2024llm}, where $N$ is the number of generated stories per instruction.

In each comparison, GPT-3.5-Turbo is prompted to provide a score out of $5$ for both stories for the criteria of grammar, coherence, and likeability. To get a normalized coherence score for the LLM given the number of constraints, we first average the scores across the user instructions and then divide the scores by $5$ (the maximum possible coherence score).

\subsection{Output Diversity}
\label{sec:perplexity}

We use two methods to measure the generation diversity of the testing LLMs: self perplexity and dist-n diversity~\citep{li2016diversity}. Self perplexities are calculated using the same LLM that generates the story. A higher self perplexity means the testing LLM also assigns probabilities to other tokens, which implies high generation diversity~\citep{hashimoto2019unifying}. Dist-n is the ratio of distinct n-grams to the total number of n-grams and we compute the products of dist-2, dist-3, and dist-4~\citep{li2022contrastive}:
\begin{equation}
\label{eq:diversity}
    \text{Dist-n Diversity} = \prod_{n=2}^{4} \frac{|\text{unique\ n-grams}|}{|\text{total\ n-grams}|}.
\end{equation}

\subsection{Creativity Measurements}

By comparing the curves of LLMs in \Cref{fig:Venn_diag} (b), we can understand that LLM2 is more creative than LLM1 since it performs better given more constraints and this performance decays slower as more constraints are introduced. To quantify such observations, we propose two new creativity metrics: Quality Under $n$ Constraints (QUC$_n$), and Relative Creativity Score (RCS$_{m,n}$), which is the quality difference in two stories generated using $m$ constraints and $n$ constraints, respectively. 

For a story generated using $n$ constraints, QUC$_n$ is defined as the product of the normalized coherence score and the average percentage of constraints satisfied. When the prompt contains many constraints (i.e., large $n$), a good output story should still be both coherent and adhere to many constraints, leading to a high to a high QUC$_n$.

To measure the quality decay speed, we define the relative creativity score RCS$_{m,n}$ as the difference between the story quality from the smallest number of constraints $m$ (QUC$_m$) and the story quality from the largest number of constraints $n$ (QUC$_n$). A smaller RCS$_{m,n}$ of an LLM indicates its slower decay speed and thus better creativity.

\section{Experiments}
\begin{figure*}[t!]
\centering
\begin{subfigure}{.49\textwidth}
  \centering
  \includegraphics[width=1\linewidth]{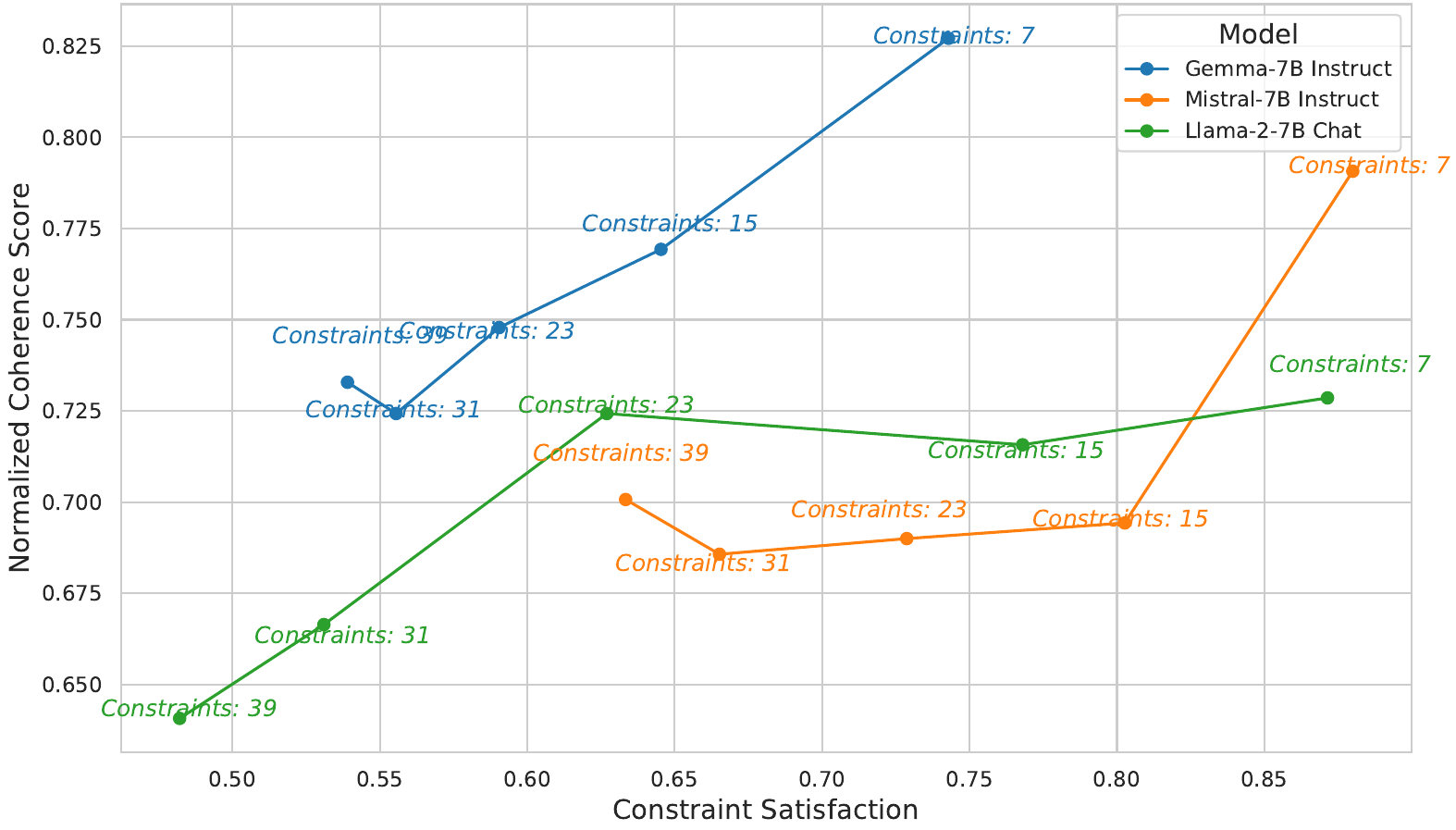}
  \caption{Story-based constraints and GPT-4 evaluator}
  \label{fig:GPT-4_Results}
\end{subfigure}%
\hfill
\begin{subfigure}{.49\textwidth}
  \centering
  \includegraphics[width=1\linewidth]{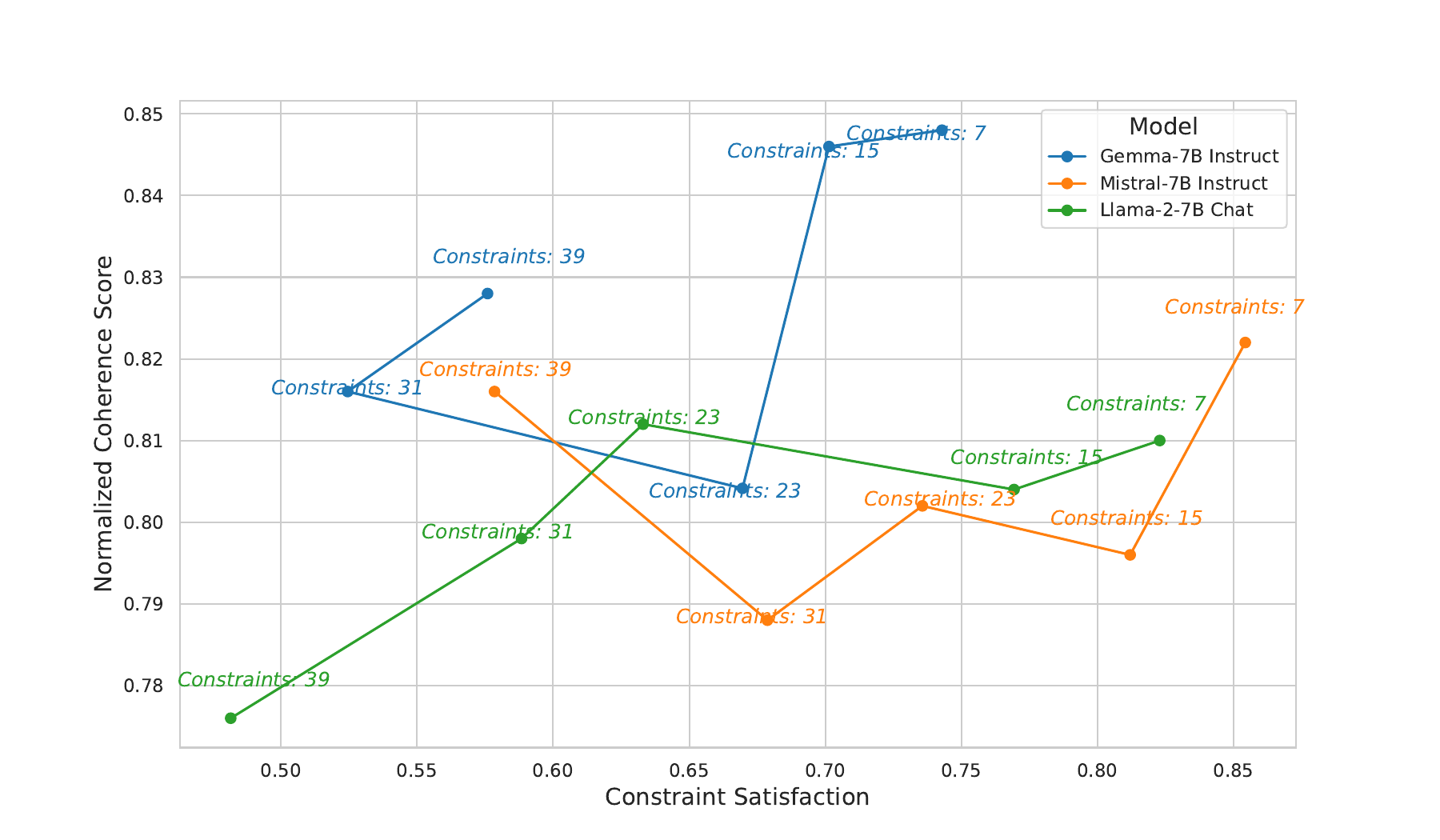}
  \caption{Story-based constraints and GPT-3.5-Turbo evaluator}
  \label{fig:Type2_tradeoff}
\end{subfigure}
\par\bigskip
\begin{subfigure}{.49\textwidth}
  \centering
  \includegraphics[width=1\linewidth]{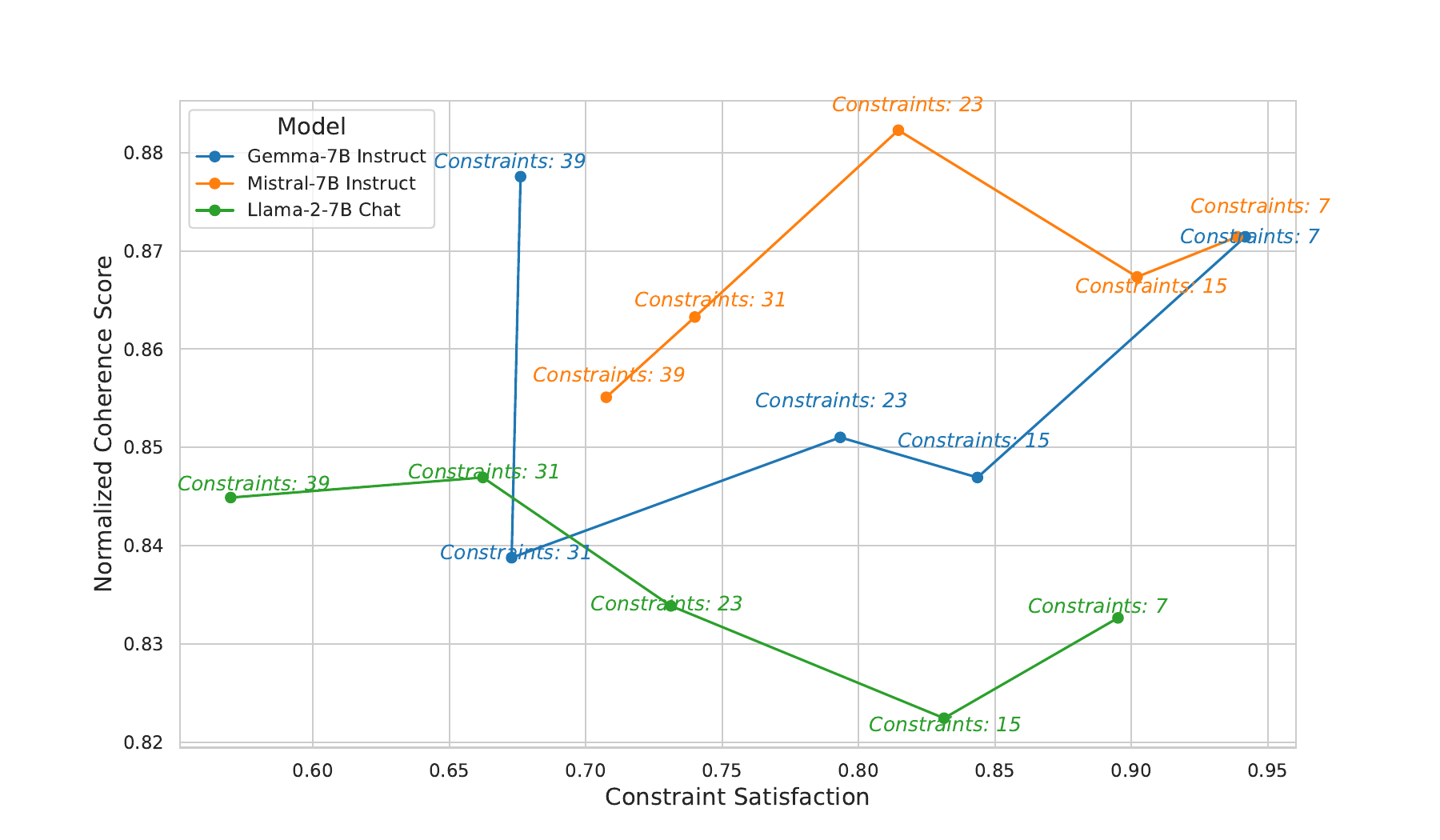}
  \caption{Instruction-based constraints and GPT-3.5-Turbo evaluator}
  \label{fig:Type1_tradeoff}
\end{subfigure}%
\hfill
% \begin{subfigure}{.45\textwidth}
%   \centering
%   \includegraphics[width=1\linewidth]{latex/Figures/Type 1OlmoResults.pdf}
%   \caption{Results on instruction-based constraints obtained using GPT-3.5-Turbo as the evaluator}
%   \label{fig:Type1_Olmo}
% \end{subfigure}%
\begin{subfigure}{.49\textwidth}
  \centering
  \includegraphics[width=1\linewidth]{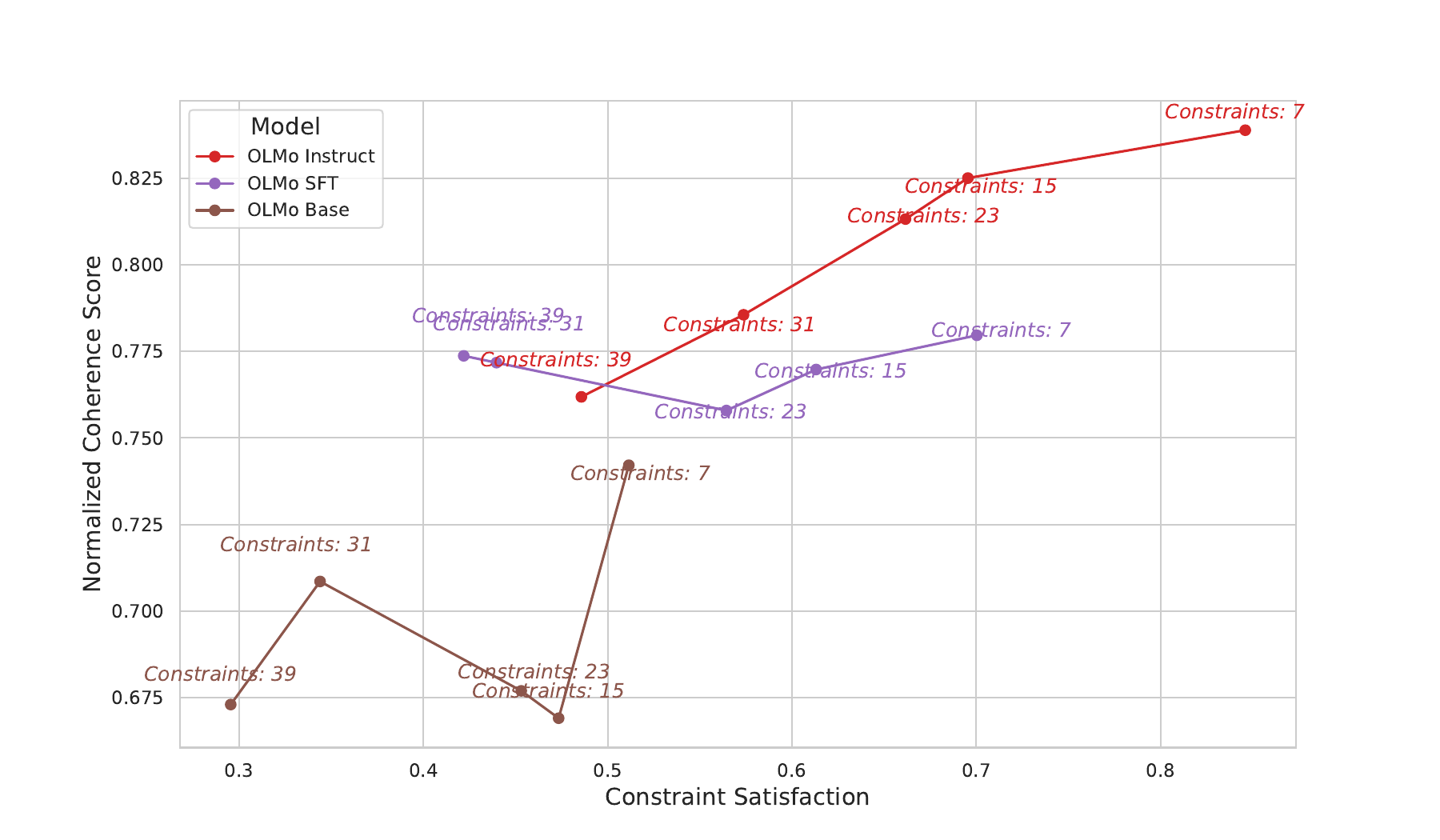}
  \caption{Story-based constraints and GPT-3.5-Turbo evaluator}
  \label{fig:Type2_Olmo}
\end{subfigure}
\caption{Analyzing the trade-off between coherence and constraint satisfaction.}
\label{fig:coh_fol_trade}
\end{figure*}

This section describes our experimental setup and analyzes the results obtained for each evaluation metric used in this study. We first highlight the trade-off between a model's instruction-following ability and the coherence of the generated stories. We then analyze the effect of LHF and SFT on text generation using OLMo, discuss our creativity measurements and the diversities of different LLMs, and identify which types of constraints are not satisfied.

%and text generation quality as prompt specificity (i.e., number of constraints) increases. 

\subsection{Setup}
In this work, we consider six models for story generation, namely LLaMA-2-7B Chat \citep{touvron2023llama}, Gemma-7B Instruct \citep{team2024gemma}, Mistral-7B Instruct \citep{jiang2023mistral}, and three versions of OLMo-7B \citep{groeneveld2024olmo} - OLMo Base, OLMo SFT (the model obtained after SFT), and OLMo Instruct (the model obtained after LHF). Each model generates stories for all the $500$ unique prompts described in \Cref{sec:constraint_num}, resulting in $3000$ stories. The temperature of the generation is $0.8$ and the threshold of top-p is $0.95$. We compare the coherence and constraint satisfaction performances of different LLMs based on the 3000 stories. 

%temperature=0.8, 
%topp=0.95

\subsection{Trade-off Between Coherence and Constraint Satisfaction}
\label{sec:Trade-off Between Coherence and Constraint Satisfaction}

If an LLM outputs the best story in its memory regardless of the prompt, it could achieve the highest coherence while having the worst constraint satisfaction. In contrast, an LLM could simply copy all the plot-related constraints into the story without proper transitions, resulting in a high constraint satisfaction score but poor coherence. Figure~\ref{fig:Trade-off} shows an example of such a trade-off. Hence it is important to evaluate both constraint satisfaction and coherence and study the interplay between them. 

To compare this trade-off in different LLMs, we plot both metrics under different numbers of constraints in \Cref{fig:coh_fol_trade}. Points at the top-right corner of the figures correspond to high coherence and constraint satisfaction scores. In \Cref{fig:GPT-4_Results}, we observe that coherence and constraint satisfaction do not decay at the same rate. Instead, different LLMs adjust the balance between coherence and satisfaction differently as the number of constraints increases. For example, \textbf{LLaMA-2} focuses on maintaining coherence by sacrificing the satisfaction probability when the number of constraints ranges between $7$ and $23$.

We use GPT-3.5-Turbo to evaluate the coherence and constraint satisfaction in the rest of the experiments because running all the experiments using GPT-4 exceeds our budget. Replacing GPT-4 in \Cref{fig:GPT-4_Results} with GPT-3.5-Turbo makes the curves in \Cref{fig:Type2_tradeoff} noisier, but the overall trends remain similar. 

Finally, we find that different constraint-synthesizing approaches could lead to different results. For example, \textbf{Gemma-7B} outputs more coherent stories in response to story-based prompts compared to \textbf{Mistral-7B} in \Cref{fig:Type2_tradeoff}, but the trend is reversed for instruction-based prompts in \Cref{fig:Type1_tradeoff}.

%Different ways to synthesize 
%comparing Figures \ref{fig:Type2_tradeoff} and 

\subsection{Effects of LHF}

A similar analysis was also conducted on \textbf{OLMo Base}, \textbf{OLMo SFT}, and \textbf{OLMo Instruct} in \Cref{fig:Type2_Olmo} to understand the impact of different stages of LLM training. Their curves are smoother compared to \Cref{fig:Type2_tradeoff}, probably because it is easier for GPT-3.5-Turbo to evaluate the stories by comparison when the stories from LLMs share similar styles and have relatively worse qualities. 

\Cref{fig:Type2_Olmo} demonstrates that \textbf{OLMo Instruct} is much better than \textbf{OLMo SFT} given only $7$ constraints, but they almost converge when they are presented with $31$ or $39$ constraints. We hypothesize that this is because LHF can help LLMs select better stories when the number of constraints is not large. For example, at \Cref{fig:Venn_diag} (b), by learning from human preferences, LLMs could select the six most coherent stories that satisfy the constraints. However, if only one relevant training story satisfies all three constraints, human preference data will not help LLMs increase the story quality or compose new stories. These insights are hard to discover in the previous benchmarks whose number of constraints is seldom greater than $10$.

\begin{table}[t!]
\centering
\scalebox{0.77}{
    \setlength{\tabcolsep}{2.65 pt} % Reduce space between columns
    \begin{tabular}{lcccc}
        \toprule
        & & & \multicolumn{1}{c}{Median} & \multicolumn{1}{c}{Self} \\
        Model       & QUC$_{39}$     &      RCS$_{7, 39}$ $\downarrow$  &            \multicolumn{1}{c}{Length} & \multicolumn{1}{c}{Perplexity}    \\
        \midrule
        Gemma-7B Instruct      & \textbf{0.4768} & \textbf{0.1531} & \textbf{488} & 3.6620\\
        Mistral-7B Chat   & 0.4720 & 0.2301 & 657 & 1.5834\\
        LLaMA-2-7B Instruct      & 0.3736 & 0.2928 & 513 & 2.2870\\
        OLMo-7B Base   & 0.1988 & 0.1807 & 531 & \textbf{3.6740}\\
        OLMo-7B SFT    & 0.3263 & 0.2195 & 572 & 3.6172\\
        OLMo-7B Instruct  & 0.3700 & 0.3395 & 843 & 3.6007\\
        \bottomrule
    \end{tabular}
    }
    \caption{Overall comparison of different LLMs on story-based constraints. Smaller RCS indicates higher creativity. Models adhering closely to the 500-word limit in our instruction show better compliance and higher self perplexities mean more diverse outputs. Both metrics are computed across all the generated stories. The best values are highlighted.}
    \label{tab:Type2metrics}
\end{table}

\subsection{Creativity Evaluation}

\Cref{tab:Type2metrics} compares the overall creativity of different LLMs. High QUC$_{n=39}$ scores for \textbf{Gemma-7B} and \textbf{Mistral-7B} indicate that these models produce stories of good quality even in the presence of several constraints. \textbf{Gemma-7B} has also achieved the lowest RCS$_{m=7, n=39}$ score indicating that it is more creative because the quality of its stories degrades slower as more constraints are introduced. 

An LLM could add more relevant plots or transitions to a longer story to satisfy more constraints or make it more coherent. Thus, although \textbf{OLMo Instruct} has a higher QUC$_{n=39}$ score than \textbf{OLMo SFT}, it is possible that the source of improvement comes from its tendency to output longer responses by ignoring the $500-$word constraint in our prompt~\citep{singhal2023long,dubois2024length}. Finally, \textbf{Gemma-7B} outputs more concise and diverse stories compared to \textbf{Mistral-7B}. Having multiple ways to satisfy many constraints in shorter stories demonstrates the creativity of \textbf{Gemma-7B}.

\begin{figure*}[t!]
\centering
\begin{subfigure}{.33\textwidth}
  \centering
  \includegraphics[width=1\linewidth]{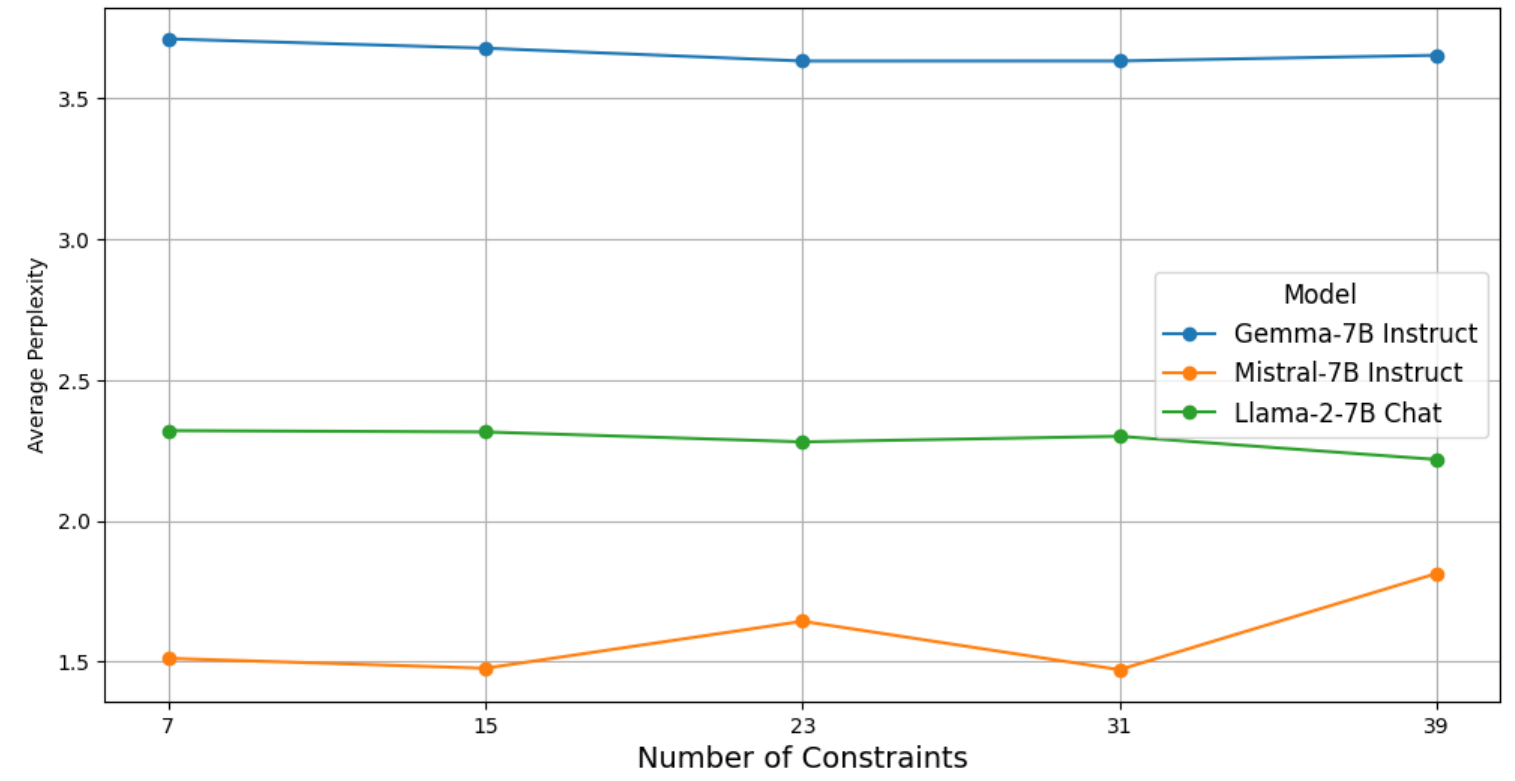}
  \caption{Self perplexity scores}
  \label{fig:D3Perplexity}
\end{subfigure}%
\begin{subfigure}{.33\textwidth}
  \centering
  \includegraphics[width=1\linewidth]{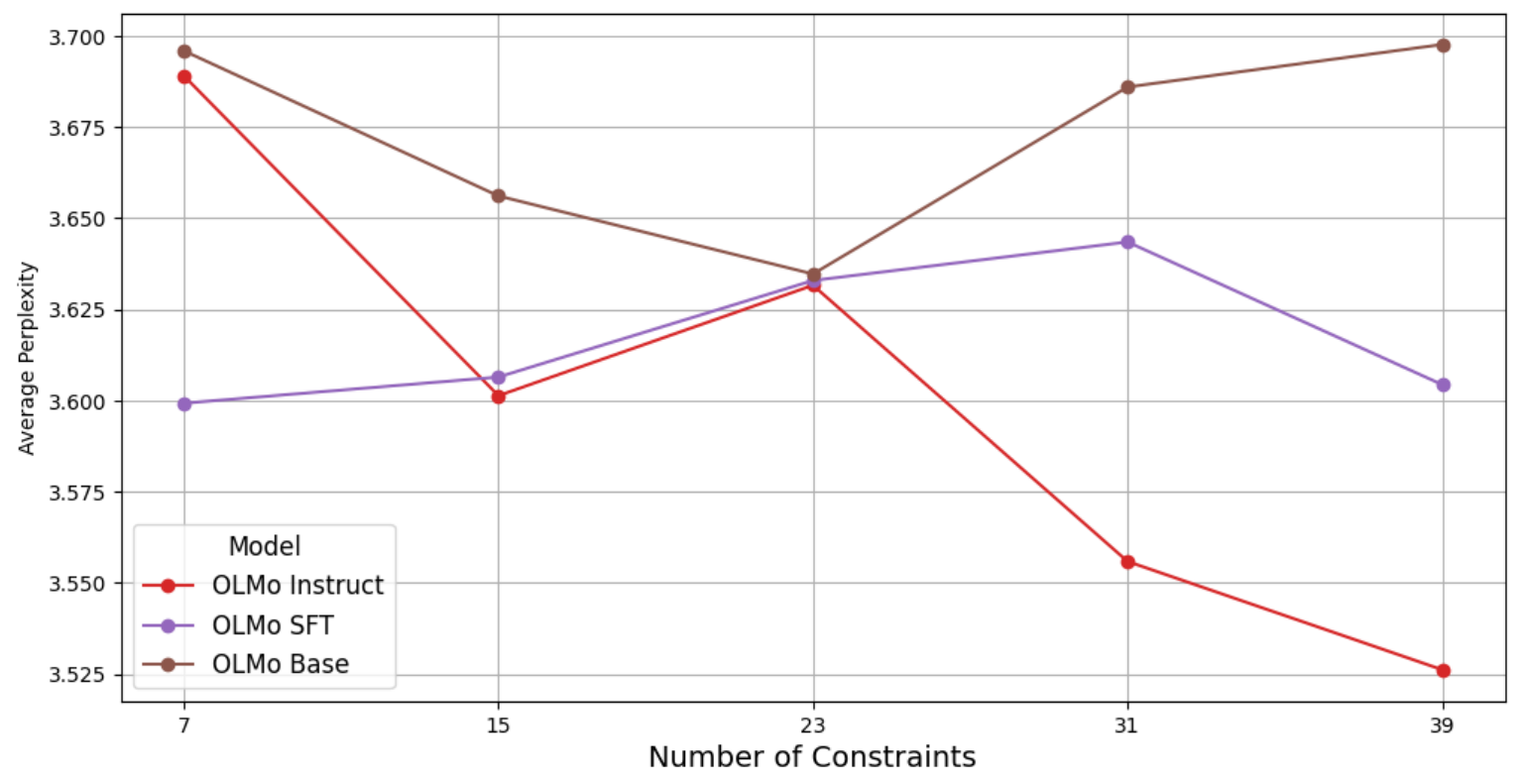}
  \caption{Self perplexity scores of OLMo series}
  \label{fig:OlmoD3Diversity}
\end{subfigure}%
\begin{subfigure}{.33\textwidth}
  \centering
  \includegraphics[width=1\linewidth]{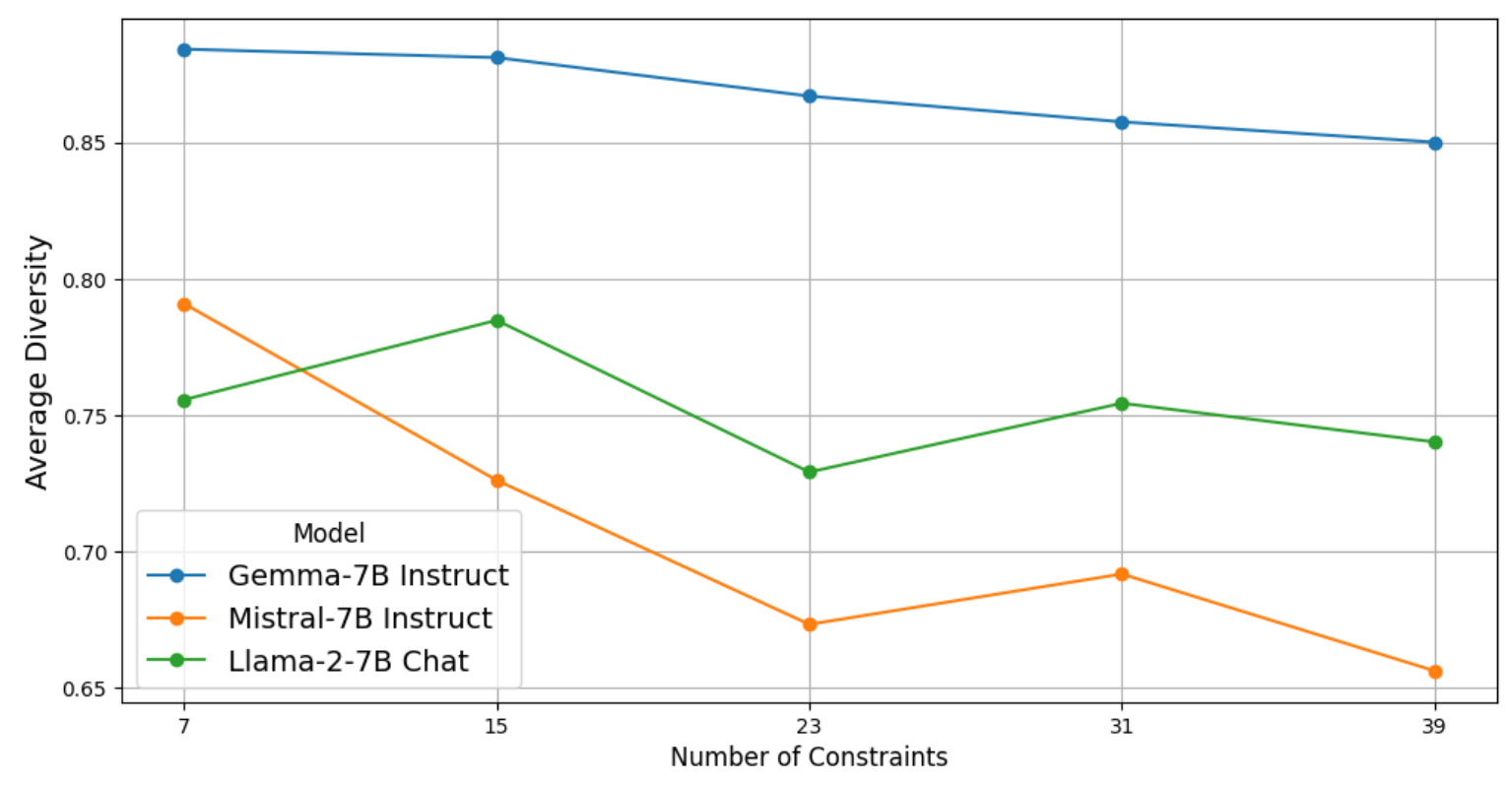}
  \caption{Dist-n diversity scores}
  \label{fig:D3Diversity}
\end{subfigure}
\caption{Generation diversity for stories written using story-based constraints.}
\label{fig:diversity_main}
\end{figure*}

%\subsection{Evaluating Story Perplexity and Diversity}
\subsection{Story Diversity Evaluation}

In \Cref{fig:OlmoD3Diversity}, \textbf{OLMo Instruct} has a slightly lower self perplexity, which is likely a consequence of LHF during model training \citep{kirk2024understanding, le2024exploring, xiao2024algorithmic}. The vocabulary diversity of \textbf{LLaMA-2-7B Chat} also decays in \Cref{fig:D3Diversity}. Except for these two exceptions, \Cref{fig:diversity_main} shows that the diversities of all LLMs do not have significant changes overall as we add more constraints into the prompt. 

These results might be caused by the balance of two opposite trends: as the number of constraints increases, LLMs have fewer relevant training examples to leverage, potentially leading to decreased diversity. On the other hand, there are more possible combinations of constraints they can choose from to satisfy, which could increase diversity. As a result, we can focus on analyzing the trade-off between coherence and constraint satisfaction, without pondering about the implications on diversity.

%they could choose to satisfy only a few constraints and generate novel stories 
%In turn,
%these results support the quality of our benchmark by allowing researchers to focus on just 

%We observed that for Llama, Mistral, and Gemma, the perplexity and diversity of the generated text had no specific relation with the number of constraints, and they neither increased nor decreased significantly with an increase in the number of constraints (Figures ~\ref{fig:D3Perplexity} and ~\ref{fig:D3Diversity}). 

%The same holds for OLMo models (Figure ~\ref{fig:OlmoD3Diversity}) as well, with the 

%To study this, we computed the perplexity and diversity of the stories as explained in Section \ref{sec:perplexity}. 

%These results indicate the LLMs do not compromise on the diversity of the generated text to maintain a high quality of the stories. 

\subsection{Error Analysis on OLMo}
\label{sec:ErrorAnal}
To delve deeper into the impact of SFT and LHF in text generation, we perform an error analysis on the responses to prompts with $39$ constraints. In the error analysis, we manually inspect around $600$ constraint satisfactions, and with the help of GPT-4, we examine all individual constraints violated by \textbf{OLMo Base}, \textbf{OLMo SFT}, and \textbf{OLMo Instruct}.
%Considering stories generated from the most complex prompts, i.e., ones with 39 constraints, we evaluated which of the OLMo models satisfied the constraints.

We observed that when the constraints were satisfied by all three models, they were usually specific actions or events about the story, or were related to the style or tone of the story. The LLMs could simply copy such constraints to the story, thus satisfying most of them. When there was a slight depth added to the character or plot, \textbf{OLMo Base} was unable to satisfy them. \textbf{OLMo Base} was also unable to satisfy inference-based constraints, for example, events reflecting societal norms in the story. Furthermore, when there was any use of obscure knowledge or interactions between different characters or plot lines, all models failed. Finally, when complex narrative patterns (like plot twists), specific roles to characters, or futuristic/imaginative concepts were given, both \textbf{OLMo Base} and \textbf{OLMo SFT} fail but \textbf{OLMo Instruct} succeeds. One possible reason could be that \textbf{OLMo Instruct} tends to output a much longer story, which has more room to satisfy more complicated constraints.

\section{Related Work}
%\input{latex/Related Work}
%To the best of our knowledge, no research work has evaluated the effect of specificity on the creative generation capabilities of LLMs. 

%stochastic parrots~\citep{bender2021dangers}

As LLMs become popular, more researchers are curious about whether LLMs could do well in tasks requiring creativity such as humor generation~\citep{zhong2024let}, comedy creation~\citep{mirowski2024robot}, generating creative analogies~\citep{kang2022augmenting,ding2023fluid,liu2024ai}, and creativity tests in psychology~\citep{bellemare2024divergent}. Previous findings suggest that LLMs could reduce the creativity of a group (i.e., homogenizing effect)~\citep{moonhomogenizing,anderson2024evaluating} while improving the creativity of individual humans~\citep{kang2022augmenting,ding2023fluid,liu2024ai}. Nevertheless, these studies do not try to measure the creativity of LLMs while preventing LLMs from memorizing something that looks creative in the training corpus.
Our goal is to provide an automatic framework for evaluating the creativity of LLMs under up to 39 constraints, which differs from the focus of previous benchmarks. For example, \citet{boussioux2023crowdless,chakrabarty2024art} recruit humans to judge if the outputs from LLMs are creative. CFBench~\citep{zhang2024cfbench} analyzes the satisfaction ratios given mostly less than 8 constraints. CELLO~\citep{he2024can},
ComplexBench~\citep{wen2024benchmarking}, and Suri~\citep{pham2024suri} provide more complex instructions without measuring LLMs' creativity using prompt specificities.

Finally, our methodology is related to the benchmarks that test LLMs' reasoning ability. For example, \citet{wu2023reasoning, nezhurina2024alice} propose reasoning tasks that are challenging for LLMs to test their limits. Another concurrent study~\citep{lu2024benchmarking} proposes to evaluate LLMs' coding creativity by prohibiting the usage of common functions such as a for-loop. Instead, our benchmark focuses on measuring story-writing creativity in practical settings.

\section{Conclusion}
In this work, we propose a novel way to measure LLMs' capability in creating original stories. The resulting \dataset benchmark has the following desired attributes, which makes it particularly useful for LLM developers. 
1) \att{Inexpensive:} We do not need human annotators to evaluate LLMs' output. For the LLMs like OLMo, the GPT3.5-Turbo evaluator is sufficient. 
2) \att{Practical:} The in-depth analysis of the benchmark reveals several actionable and undiscovered insights. For example, prioritization of coherence and instruction-following shifts for different LLMs, types of prompts, and prompt specificities; LHF's effectiveness diminishes severely in the presence of more constraints.
%in-depth analysis reveal many actionable insights that could not be discovered using only a few constraints. 
%LHF does not improve the creativity
3) \att{Simple:} By simply synthesizing more constraints and designing proper domain-specific prompts, we can measure the creativity of LLMs. The simplicity makes our methodology applicable to many domains other than story generation.

%flexible
%You can extend our method to other domains by adding more constraints and measure the performance changes.

%These attributes are particularly useful for LLM developers. 

% that are hard to be discovered without varying prompt specificities

\section{Acknowledgement}

This work was supported in part by Amazon, in part by Center for Data Science, and in part by the Center for Intelligent Information Retrieval. Any opinions, findings and conclusions or recommendations expressed in this material are those of the authors and do not necessarily reflect those of the sponsor.

%\section{Limitations and Future Directions}
\section{Limitations}
%Exploring the potential of LLMs in creative and technical fields presents numerous exciting avenues for future research. 

%We acknowledge further limitations of the current work. 

As prompts become more specific, tasks could also become difficult for humans~\citep{tulving1973encoding}. Since we haven’t conducted human experiments, we are unable to compare LLMs' performances with human performances as in \citet{west2023generative,tian2024large}. Next, professional writers have diverse ways to leverage LLMs~\citep{ippolito2022creative}, we do not know how representative our prompts in \dataset are. Furthermore, while LLM-as-the-Judge is supported and widely adopted by previous research \citep{chiang2023closer,gilardi2023chatgpt,jiang2023followbench,qin2024infobench,chia2024instructeval,zhang2024cfbench,wen2024benchmarking}, we would like to perform a human evaluation of the generated stories for their quality and compare the correlation between the evaluation judgment of LLMs and humans.

Although we can run the LLM evaluators automatically, the dataset construction still requires several iterations of prompt refinements. If we want to measure the creativities for more applications, automating the prompt design and prompt difficulty control is an interesting future project. To achieve the goal, it might be necessary to measure the specificity/difficulty of each constraint and automatically detect conflicts between constraints. Moreover, we haven't investigated if various agent-like or sophisticated prompting strategies such as \citet{zhong2024let,zhao2023self,mirowski2024robot,wei2024chain} could boost the creativity in \datasetnospace.

Finally, there are many different ways to define and measure creativity. LLMs could do very well in psychological creativity tests \citep{bellemare2024divergent}, but not so well in other measurements such as comedy creation \citep{bellemare2024divergent}. We only adopt one creativity definition (i.e., outputting original contents) and provide one inexpensive yet indirect way to measure the creativity of LLMs. The creativity measurements are tangled with other abilities of LLMs. For example, an LLM could be good at outputting an original story but bad at following instructions. \dataset currently may not be able to assign a high creativity score on such an LLM. In this case, directly measuring the similarity between the output story and the stories in the training data might be expensive but necessary.

\section{Impact Statement}
Creativity is a signature of human intelligence, so measuring the creativity gap between humans and AI models could help us predict if or when AI models could achieve artificial general intelligence. By meticulously assessing the impact of prompt specificity, our research can advance our understanding of the gap between artificial creativity and human-like storytelling and potentially be extended to measure creativity in more domains.

Other researchers could extend our conclusions, potentially leading to negative impacts on our research. For example, they might use our observations as evidence to argue that an LLM is merely a parrot without reasoning abilities, or that LHF is not an effective approach.

\bibliography{custom}

\newpage

\appendix
\section{Appendix}
\begin{figure*}[th!]
  \includegraphics[width=1\linewidth]{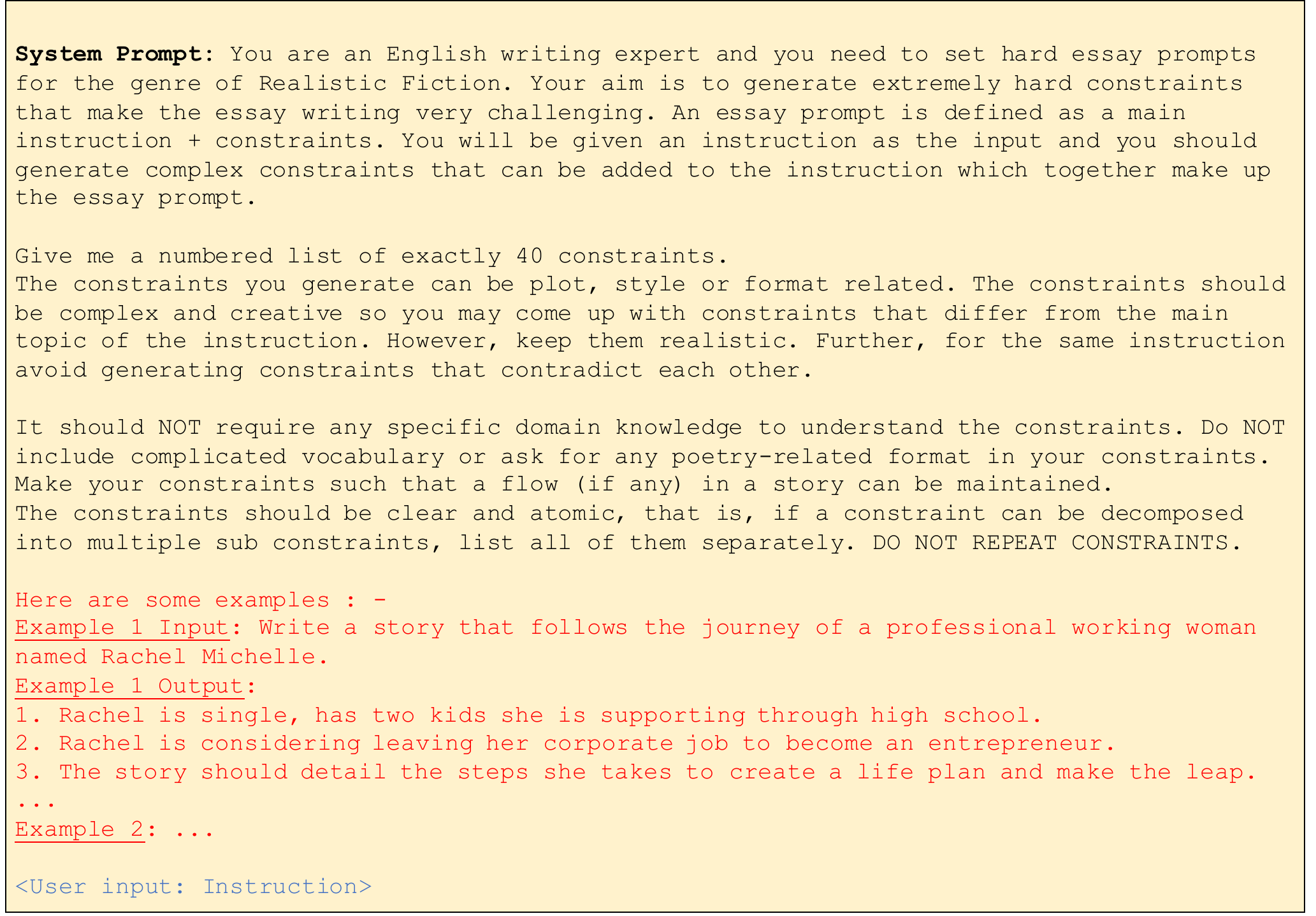}
  \caption{General framework of the system prompt used to generate instruction-based constraints.}
  \label{fig:D2 Prompt}
\end{figure*}

\label{sec:appendix}
In the supplementary material, we describe more details about our methods and prompting strategy, present some examples from our \dataset benchmark, and discuss more experimental results, focusing on those obtained from instruction-based constraints.

\subsection{Prompting Strategy for Constraints Generation}
\label{sec:ConstraintGenPrompts}
Creating the \dataset benchmark and evaluating different stories generated using the benchmark depends largely on the prompts given to the LLMs. Here, we detail the various carefully crafted prompts that primarily use few-shot prompting, reasoning, and LLM-as-the-judge prompting strategies. 
\citet{qin2024infobench} noted that on average, prompts with over 6.29 constraints (or requirements, as they call it), are hard for LLMs to satisfy comprehensively. To challenge LLMs in creative story writing, we thus start by providing LLMs with prompts containing 7 constraints, and in our \dataset benchmark, we progressively increase the number of constraints by 8, obtaining prompts with $7$, $15$, $23$, $31$, and $39$ constraints. 
As elaborated in \Cref{sec: Dataset Creation}, we use two strategies for generating constraints, namely instruction-based and story-based constraints generation. 
While generating constraints solely based on an input instruction given to GPT-4, controlling the diversity of the generated constraints is crucial. Given no restrictions, GPT-4 generates constraints corresponding to different story genres for the same instruction. Such constraints depreciate the quality of the benchmark since it makes story writing extremely difficult. To overcome this, we restricted the genre of constraints to Realistic Fiction. For the same purpose, we prompted the model to avoid generating constraints that contradict one another. We also prompted the model to avoid generating constraints that require specific domain knowledge or using poetry-related terminology. Having such constraints would highly curb a model's ability to write creative, original stories. Finally, we prompted the model to come up with atomic constraints \citep{min2023factscore} to 1) ensure that prompts with more constraints are more specific and complex than those with fewer constraints, and 2) make evaluating stories generated using \dataset easy. We provided few-shot examples illustrating each of these requirements. \Cref{fig:D2 Prompt} provides a general framework of the prompt used to develop instruction-based constraints. 

\begin{figure*}[th!]
  \includegraphics[width=1\linewidth]{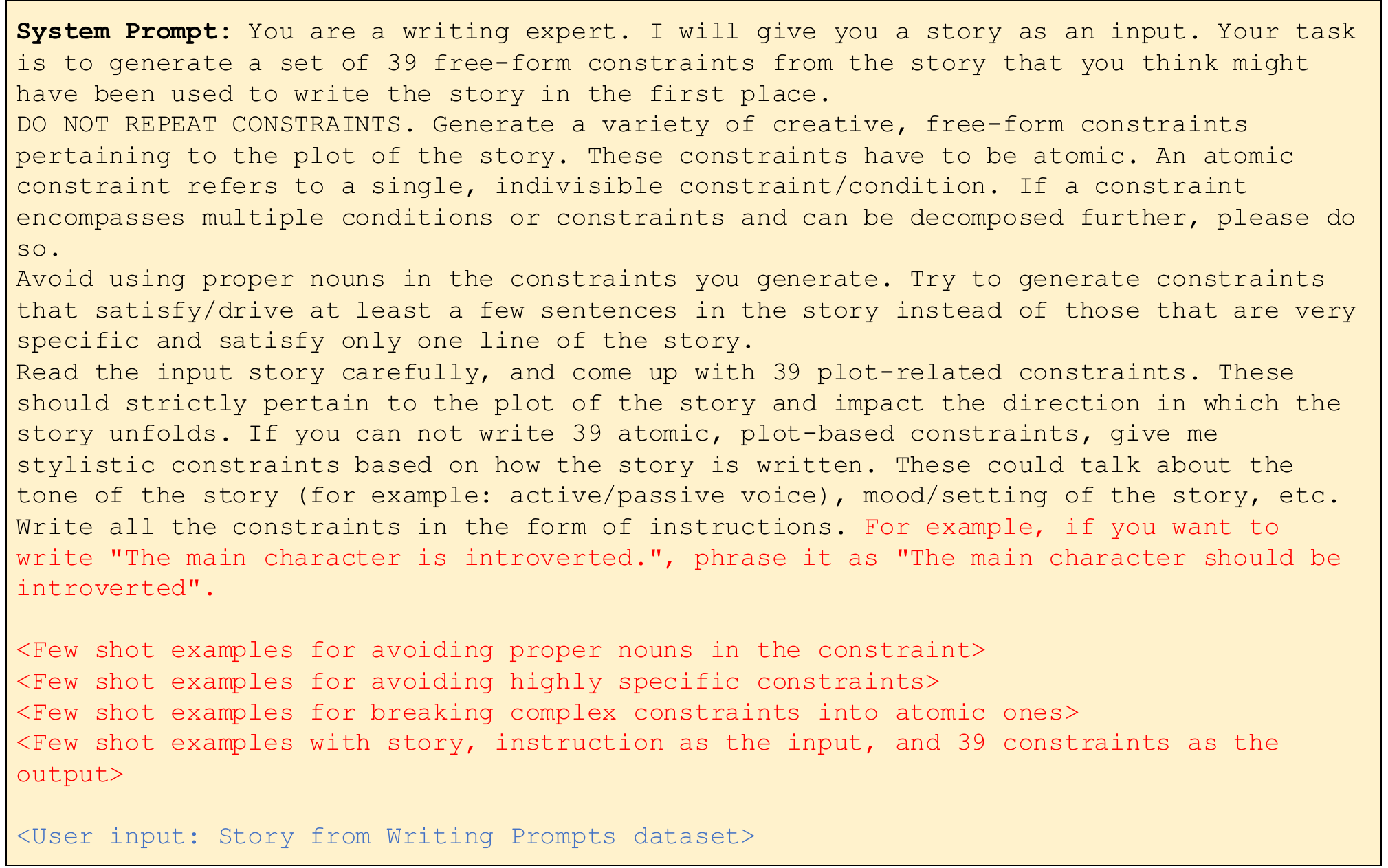}
  \caption{General framework of the system prompt used to generate story-based constraints.}
  \label{fig:D3 Prompt}
\end{figure*}

\begin{figure*}[th!]
  \includegraphics[width=1\linewidth]{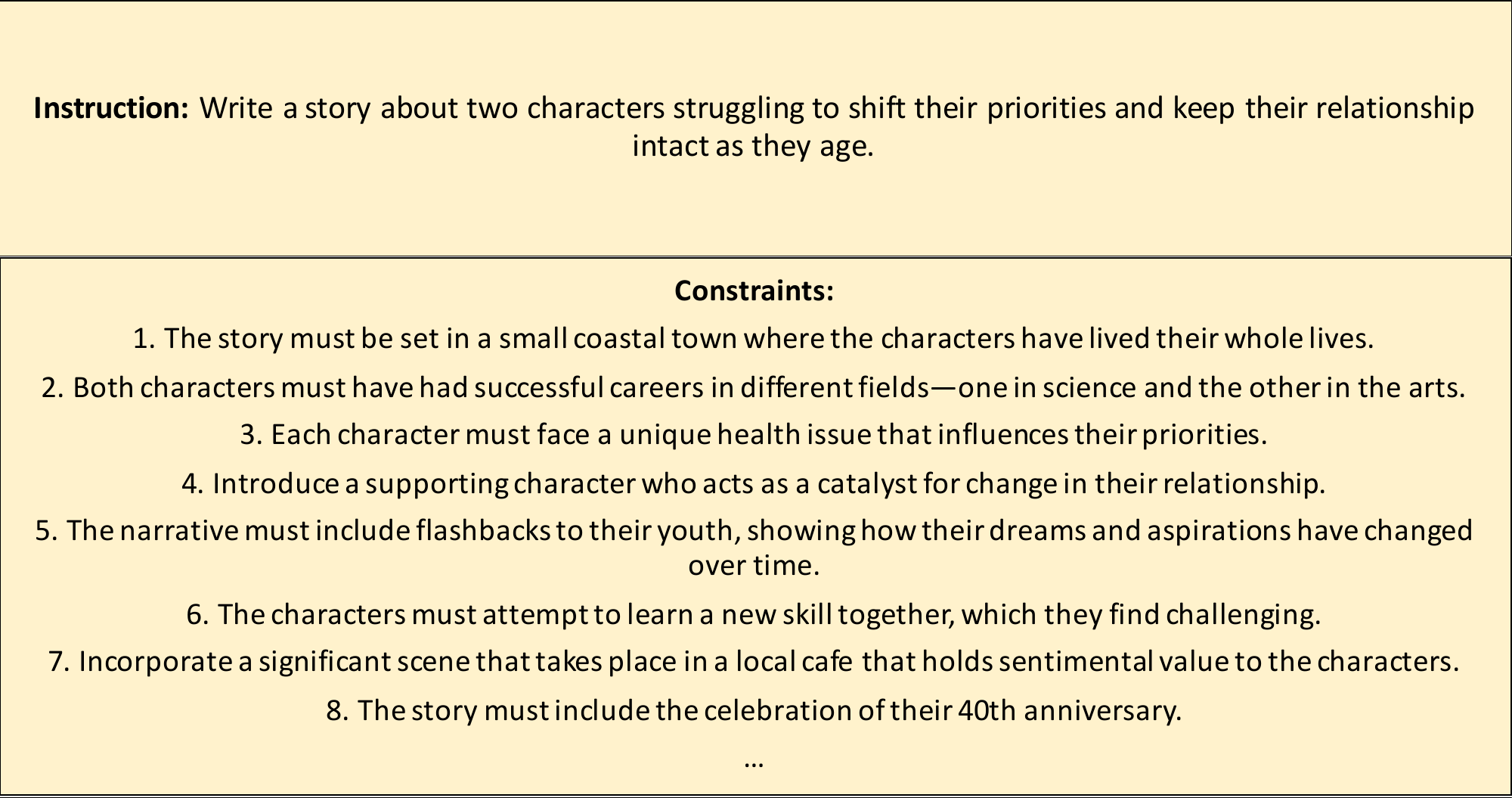}
  \caption{Example of instruction-based constraints along with the corresponding instruction.}
  \label{fig:D2 Constraints}
\end{figure*}

\begin{figure*}[th!]
  \includegraphics[width=1\linewidth]{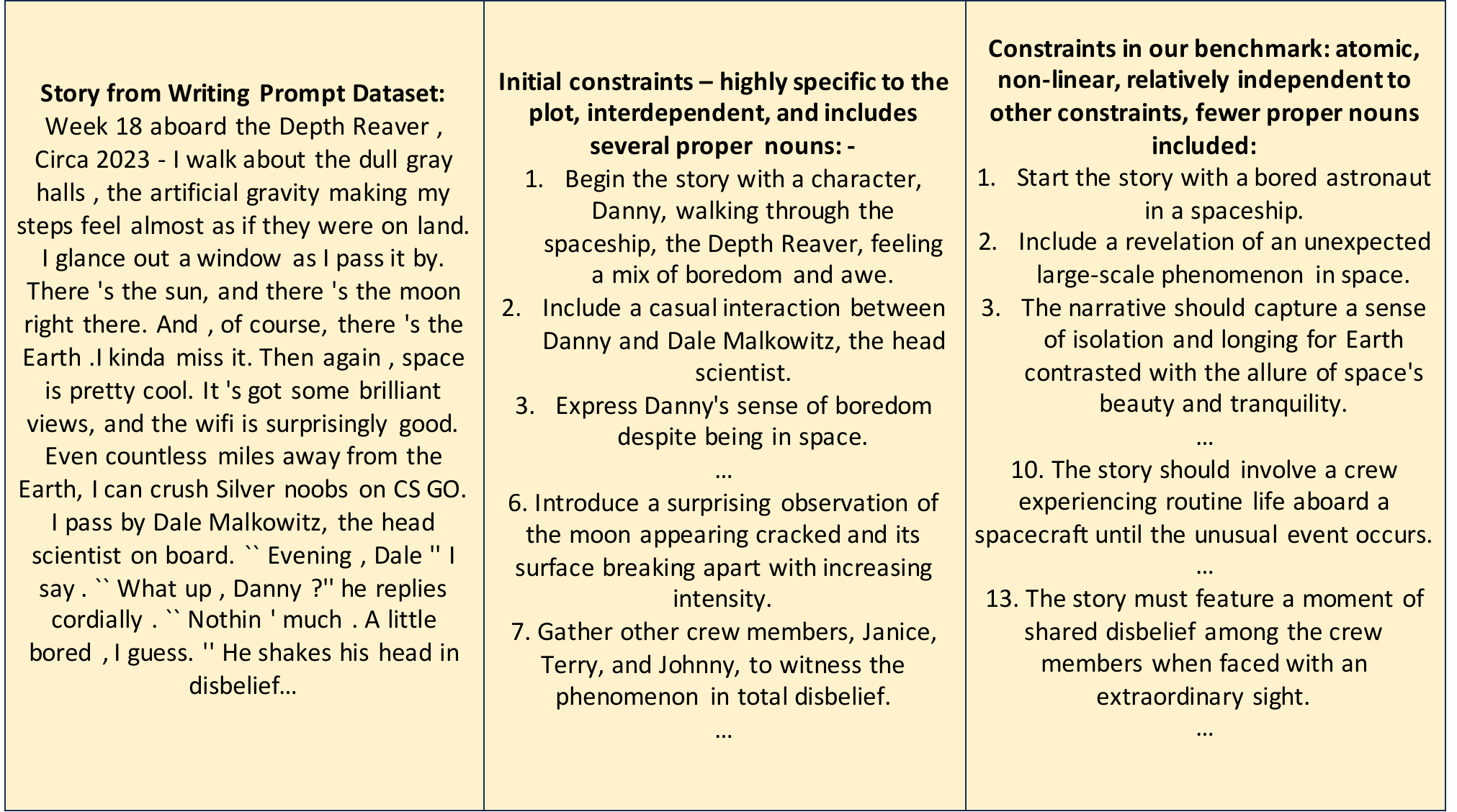}
  \caption{Example of story-based constraints before and after accounting for atomicity, linearity, specificity, and proper nouns.}
  \label{fig:Type2:OldVNew}
\end{figure*}

While generating story-based constraints, some of the above issues become irrelevant since the model generates constraints based on a story that already exists. For example, it is not possible to generate contradicting constraints or constraints about different genres from the same story. However, it is more likely that the model incorporates different proper nouns from the story into the constraints, copies specific events from the plot into the constraints, and maintains the flow of the plot in the constraints. All this would allow any LLM or story writer that uses the \dataset benchmark to copy the constraints one after another to generate a meaningful, coherent story. This could be considered ``cheating'' and would not allow one to appropriately evaluate the model for its creativity and coherence. To handle these issues, we prompted GPT-4 to generate constraints avoiding proper nouns and encouraging it to generate constraints that are non-linear with respect to the plot of the story and satisfy multiple lines of the plot rather than a highly specific event. Finally, to generate $39$ diverse constraints, it is crucial to have stories that are long enough. If not, the model would start repeating constraints or generate more constraints that are style-related, than plot-related. To overcome this, we sorted the stories from the Writing Prompts dataset in decreasing order of length and picked stories that did not violate any of GPT-4's human-alignment policies. We provided few-shot examples illustrating each of these requirements. \Cref{fig:D3 Prompt} provides a general framework of the prompt used to develop story-based constraints. Figures~\ref{fig:D2 Constraints} and \ref{fig:Type2:OldVNew} show some examples of constraints generated using the instruction-based and story-based strategies, respectively. 

\subsection{Prompting Strategy for Story Generation}
\begin{figure*}[th!]
  \includegraphics[width=1\linewidth]{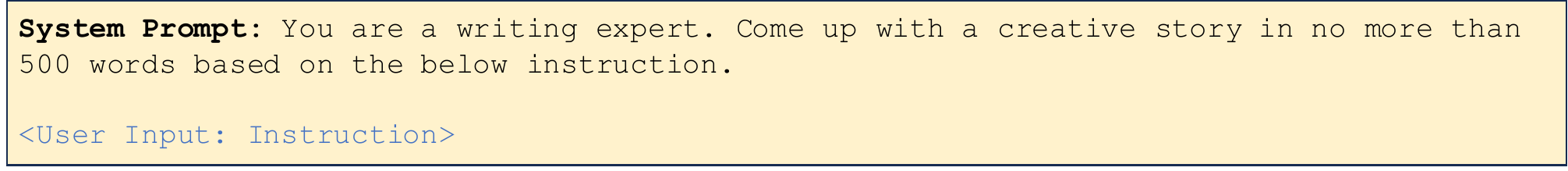}
  \caption{General framework of the system prompt used to generate the base story using GPT-4.}
  \label{fig: Base Story Generation Prompt}
\end{figure*}

\begin{figure*}[th!]
  \includegraphics[width=1\linewidth]{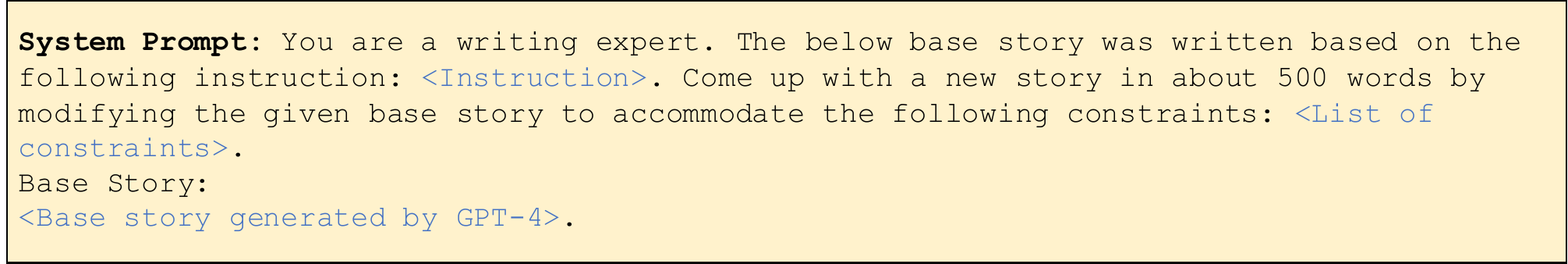}
  \caption{General framework of the system prompt used to modify the GPT-4-generated base story to accommodate different constraints.}
  \label{fig: Base Story Modification Prompt}
\end{figure*}

Compared to the prompts for generating constraints in \Cref{sec:ConstraintGenPrompts}, the prompts for story generation are straightforward. First, we prompt GPT-4 to generate a ``base'' story providing just the instruction as the input (\Cref{fig: Base Story Generation Prompt}). This is done by setting the parameters of temperature, top\_p (sampling strategy), and max\_tokens to $0.8$, $0.95$, and $4096$, respectively. The six 7B models are next tasked with modifying these base stories to satisfy different constraints (and hence prompts of different specificities) in \dataset (\Cref{fig: Base Story Modification Prompt}). All stories are generated using the NVIDIA superpod-a100 GPU consisting of $12$ nodes, and VLLM for efficient resource utilization, faster inference, and scalability. 

\subsection{Prompting Strategy for Evaluating Constraint Satisfaction} 

\begin{figure*}[th!]
  \includegraphics[width=1\linewidth]{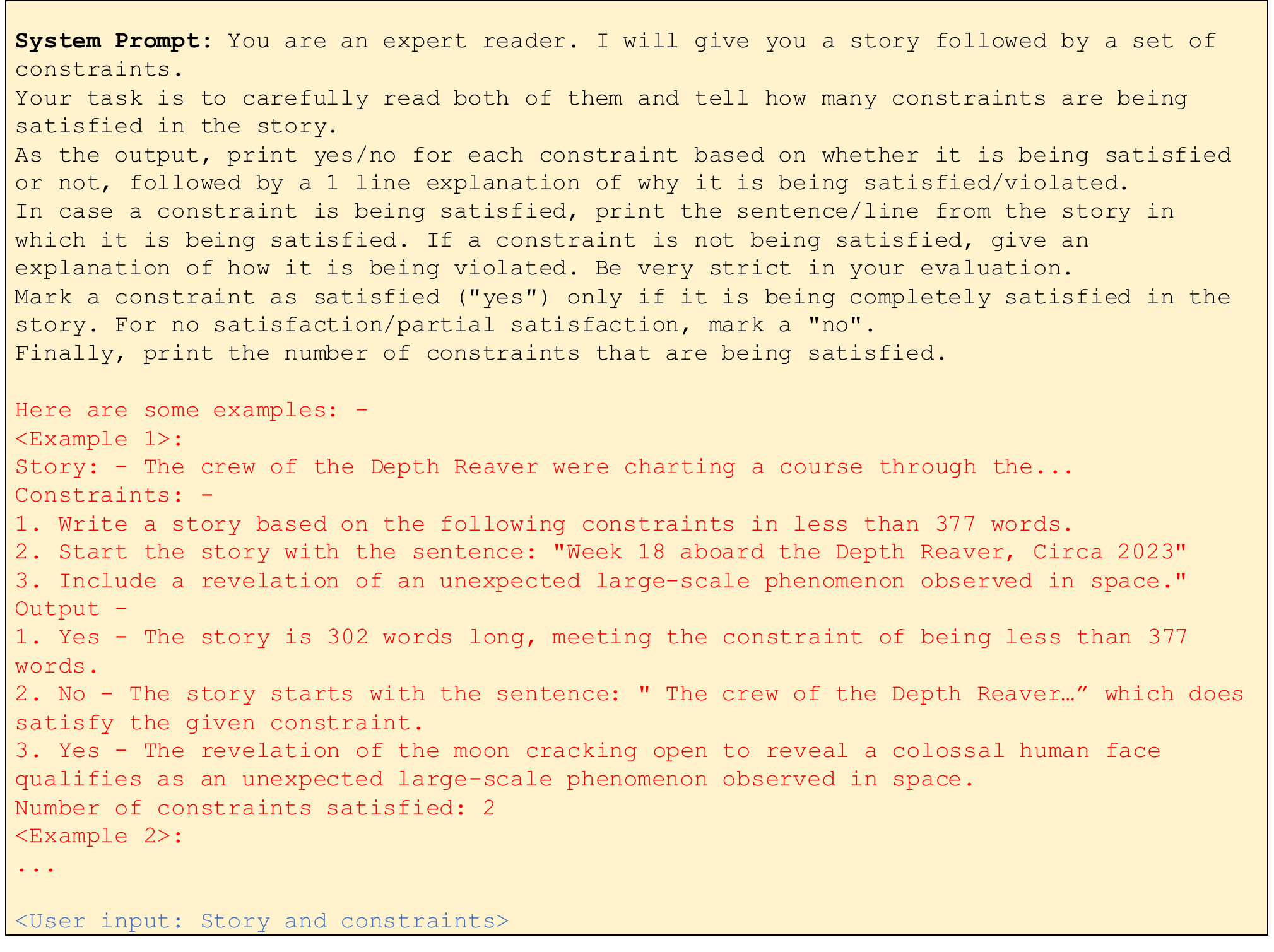}
  \caption{General framework of the system prompt used to evaluate stories for constraint satisfaction.}
  \label{fig: Constraint Satisfaction Eval Prompt}
\end{figure*}

\Cref{fig: Constraint Satisfaction Eval Prompt} shows the framework of the prompt used to evaluate the stories for their constraints satisfaction. \citet{qin2024infobench} suggests breaking down prompts into simpler criteria that can be easily evaluated to understand a model's instruction following ability. Inspired by this framework, we use a reasoning-based approach to compute the number of constraints satisfied by each story. For each constraint, we ask the evaluator model to reason if a constraint is not satisfied. On the other hand, if a constraint is satisfied, we ask the evaluator model to print the sentences of the stories that adhere to the constraint. Finally, we prompt the model to print the total number of constraints the story satisfies. 
 
\subsection{Prompting Strategy for Evaluating Coherence and Story Quality}
\begin{figure*}[th!]
  \includegraphics[width=1\linewidth]{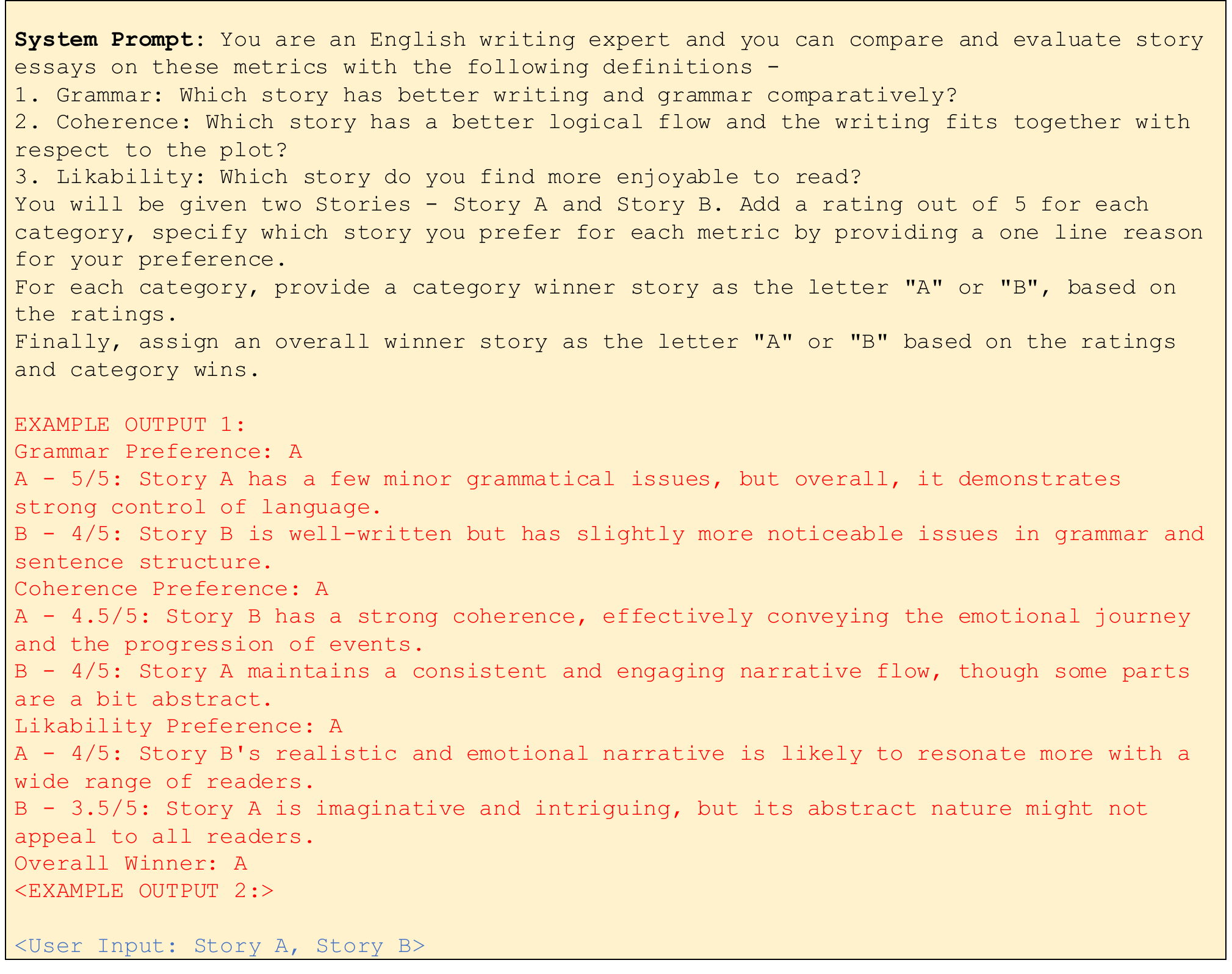}
  \caption{General framework of the system prompt used to compare two stories in terms of coherence, grammar, and likability.}
  \label{fig: Coherence Eval Prompt}
\end{figure*}

\begin{figure}[th!]
  \includegraphics[width=\columnwidth]{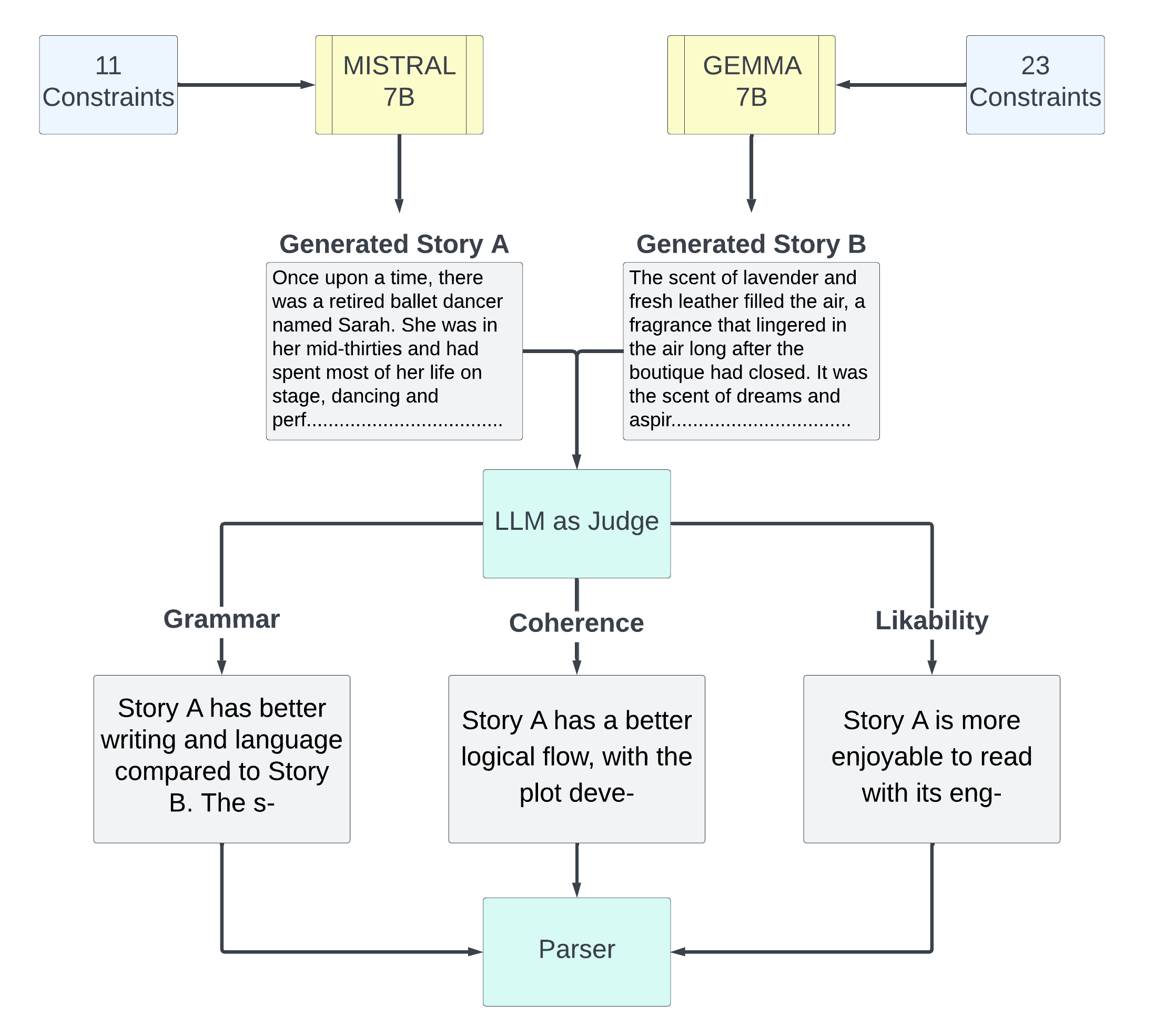}
  \caption{Example of a single pair-wise comparison for evaluating story quality.}
  \label{fig:Pair-wise}
\end{figure}

To evaluate the stories for their quality, we develop a comparative framework that compares the quality of a story in terms of likeability, grammar, and coherence, with another story. We use LLM-as-the-judge, prompting GPT-4 to provide a score out of $5$ for the two stories under consideration based on likeability, grammar, and coherence, along with a reason for the scores (\Cref{fig: Coherence Eval Prompt}). We then ask the model to pick a winner between the two stories for each category, and an overall winner based on all three categories. The obtained coherence scores are then normalized and used to plot the trade-off curves in \Cref{sec:Trade-off Between Coherence and Constraint Satisfaction}. An example of such a pair-wise comparison is shown in \Cref{fig:Pair-wise}.

However, we do not compare every unique pair of stories. Instead, as described in \Cref{sec:Story Coherence and the Linear Evaluation Algorithm} we pick a story written using $23$ constraints as the middle story with which all other stories are compared. We don't compare the middle story with itself. The story set as the middle story is compared $14$ more times than any other story. So, we instead divide its coherence score by $14$ to provide a fair comparison and normalized coherence score.

%HS moved the content from the main paper
\section{Evaluation Details}

\begin{figure}[th!]
  \includegraphics[width=\columnwidth]{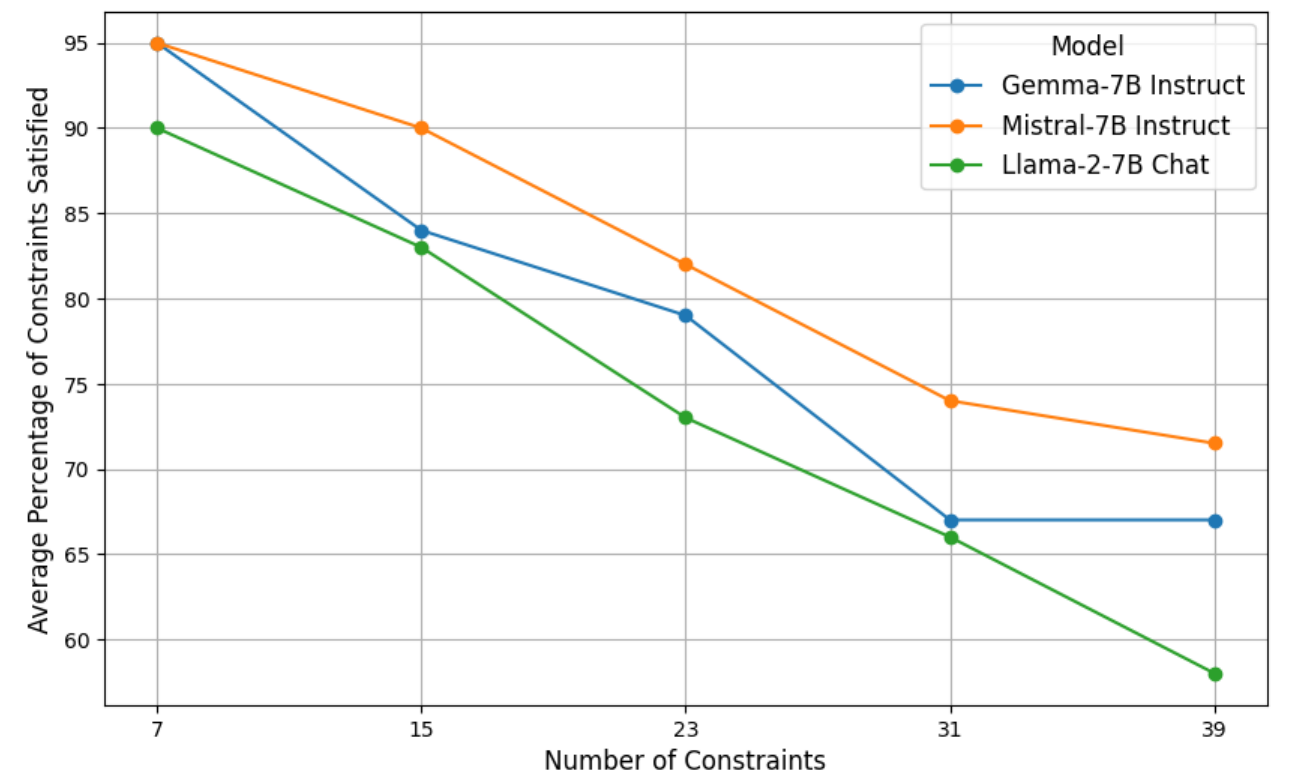}
  \caption{Percentage of constraints satisfied by storied generated by different LLMs.}
  \label{fig:ConstraintSatisfaction}
\end{figure}

This section briefly discusses additional results obtained upon evaluating the stories generated using the \dataset benchmark, focusing primarily on those obtained from instruction-based constraints. First, as the number of constraints increases, the LLMs find it harder to satisfy all of them while coming up with the stories. This essentially means that the model's instruction-following ability decreases as the specificity of the prompt increases. \Cref{fig:ConstraintSatisfaction} illustrates this result by measuring the average percentage of constraints satisfied across all stories generated from the \dataset benchmark in this work. It shows that the percentage of constraints satisfied by the models decreases with an increase in the number of constraints, i.e., an increase in prompt specificity. In this paper, GPT-3.5-Turbo refers to \textit{gpt-3.5-turbo-0125}.

\begin{figure}[th!]
  \includegraphics[width=1\linewidth]{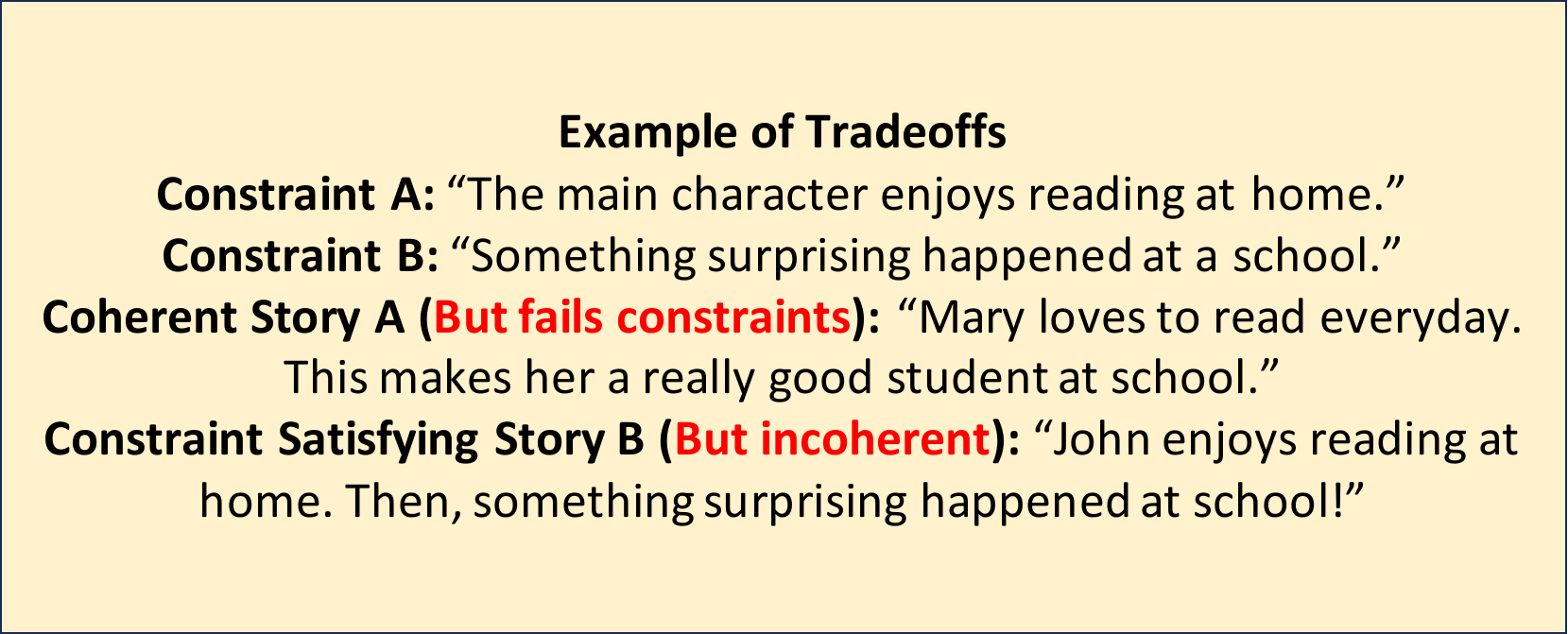}
  \caption{An example of a trade-off between constraint satisfaction and coherence.}
  \label{fig:Trade-off}
\end{figure}

\begin{figure}[th!]
  \includegraphics[width=1\linewidth]{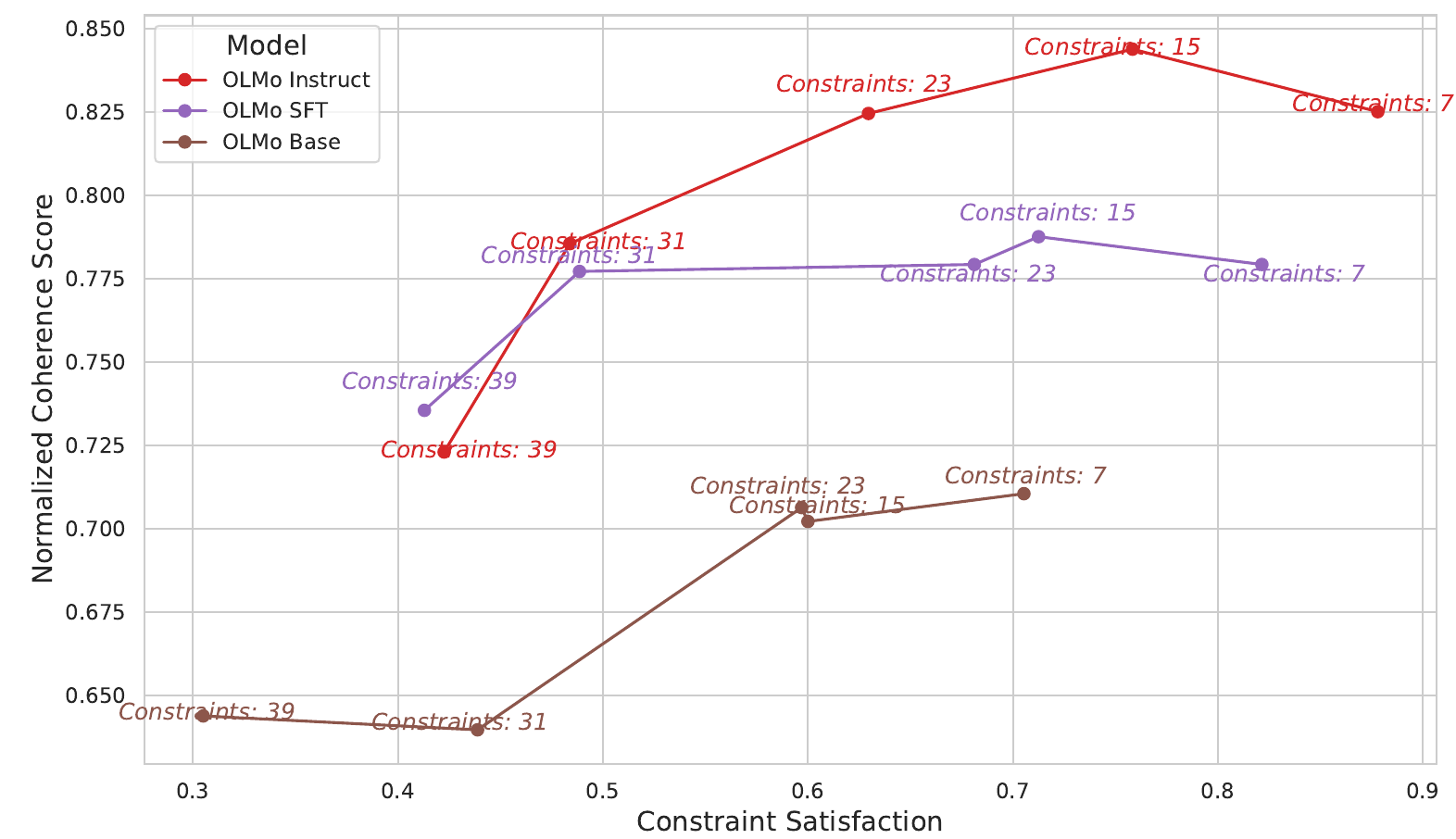}
  \caption{Trade-off results for OLMo Base, OLMo SFT, and OLMo Instruct models for instruction-based constraints.}
  \label{fig:Type1_Olmo}
\end{figure}

Additionally, as discussed in \Cref{sec:Evaluation Metrics}, evaluating the stories for both instruction-following and quality is crucial. This is because models tend to compromise one for the other when prompt specificity increases. \Cref{fig:Trade-off} shows a toy example of such a trade-off. Figures~\ref{fig:Type1_tradeoff} and \ref{fig:Type1_Olmo} depict this trade-off quantitatively for stories generated using instruction-based constraints. While it can be observed that these stories satisfy more constraints compared to those generated from story-based constraints because of the inherent nature of the generation process, the trade-off can still be observed clearly. \Cref{fig:Type2_Olmo} and \ref{fig:Type1_Olmo} further illustrate that while LHF is helpful in the presence of fewer constraints, its impact in leveraging useful data in the presence of several constraints decreases. The convergence of OLMo SFT and OLMo Instruct curves as the constraints become 31 or 39 evidence this. 

\begin{figure}[t]
  \includegraphics[width=\columnwidth]{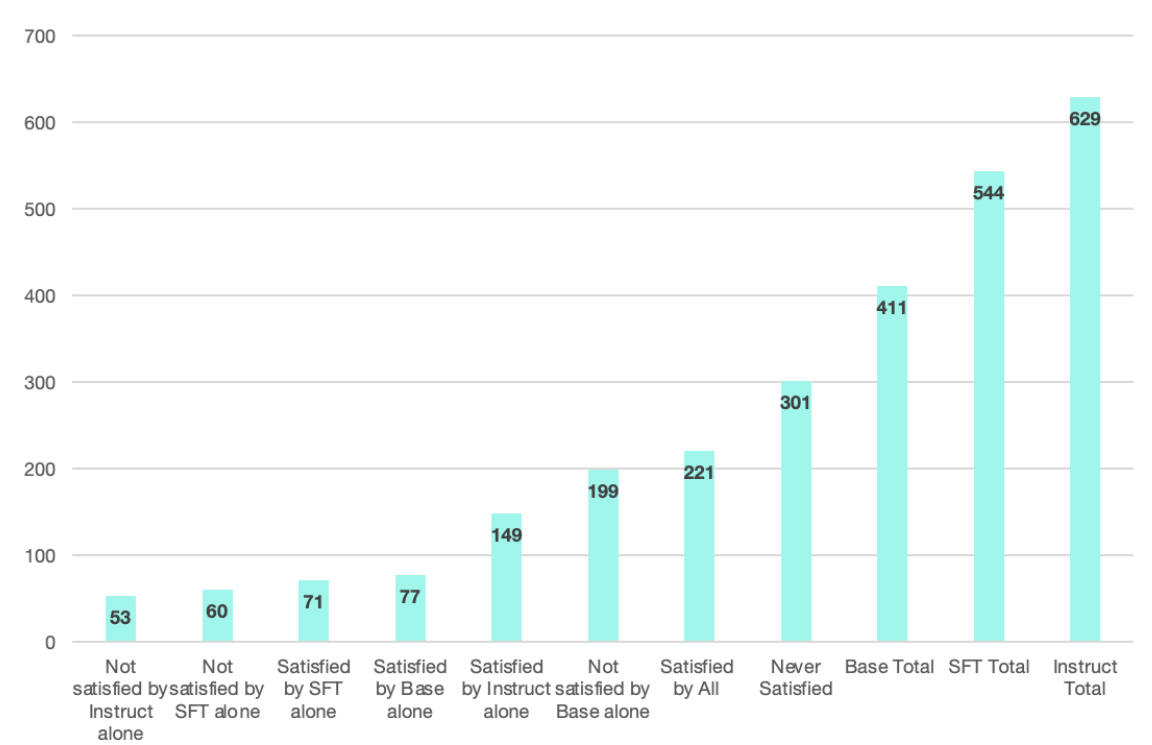}
  \caption{Error analysis for stories developed using story-based prompts with $39$ constraints.}
  \label{fig:ErrorAnal}
\end{figure}

Next, to perform error analysis, we considered all stories generated by the three OLMo models for prompts with $39$ constraints. These represent the most complex inputs in \dataset. From the $2$ constraint generation strategies, $50$ instructions in each, and $3$ OLMo models, we obtain a total of $300$ stories for evaluation. We manually inspected all stories corresponding to $5$ different instructions for their constraint satisfaction, resulting in a manual inspection of $585$ constraints ($5$ instructions $*$ $39$ constraints $*$ stories from $3$ OLMo models). We then uploaded all $300$ stories, their corresponding constraint satisfaction results, and our observations from manual inspection to the OpenAI website asking GPT-4 to refine our findings with some examples. After verifying the examples, we concluded our analysis in \Cref{sec:ErrorAnal}. \Cref{fig:ErrorAnal} illustrates these results. It depicts the number of constraints satisfied by different combinations of OLMo models, with the Instruct model satisfying the greatest number of constraints. 

\begin{figure*}[h]
    \centering
    \begin{subfigure}{0.49\textwidth}
        \centering
        \includegraphics[width=\textwidth]{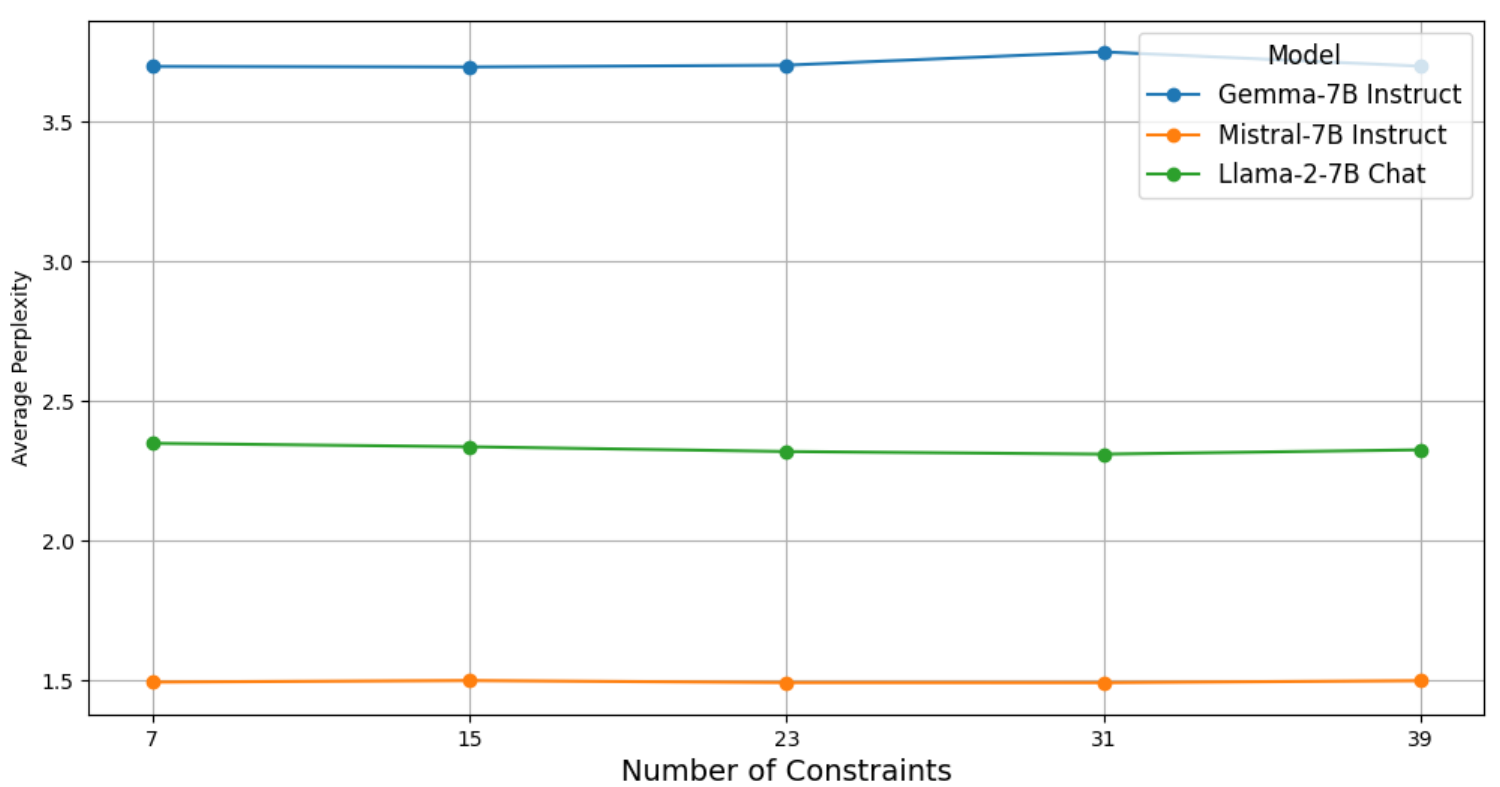}
        \caption{Perplexity scores for Mistral, Gemma, and LLaMa-2 on instruction-based constraints.}
        \label{fig:Mistral_D2Perplexity}
    \end{subfigure}%
    \hfill
    \begin{subfigure}{0.49\textwidth}
        \centering
        \includegraphics[width=\textwidth]{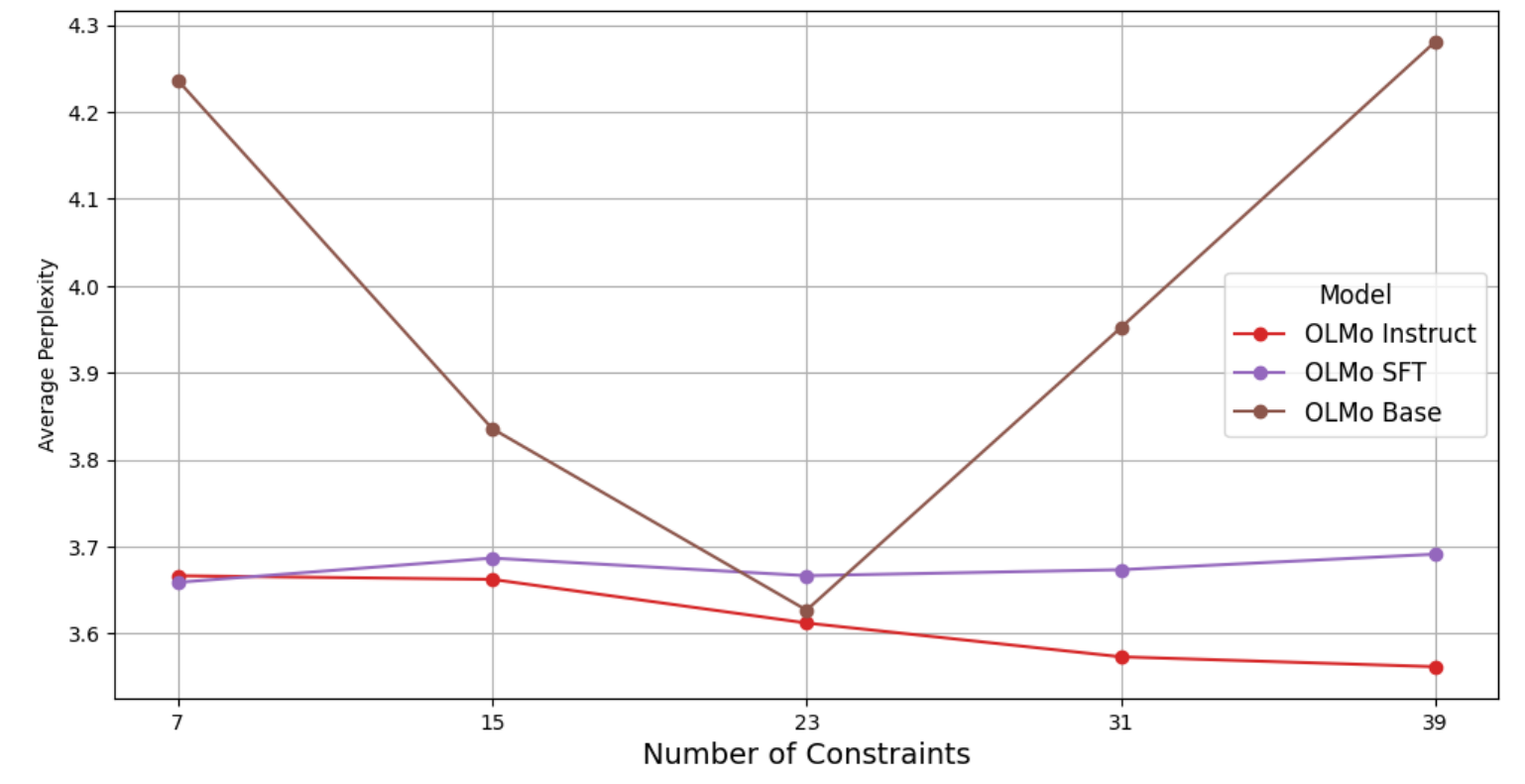}
        \caption{Perplexity scores for OLMo Base, SFT, and Instruct on instruction-based constraints.}
        \label{fig:OlmoD2Diversity}
    \end{subfigure}
    
    \caption{Perplexity scores for stories developed using instruction-based constraints.}
    \label{fig:Type1_Perplexity_combined}
\end{figure*}

\begin{figure}[th!]
  \includegraphics[width=\columnwidth]{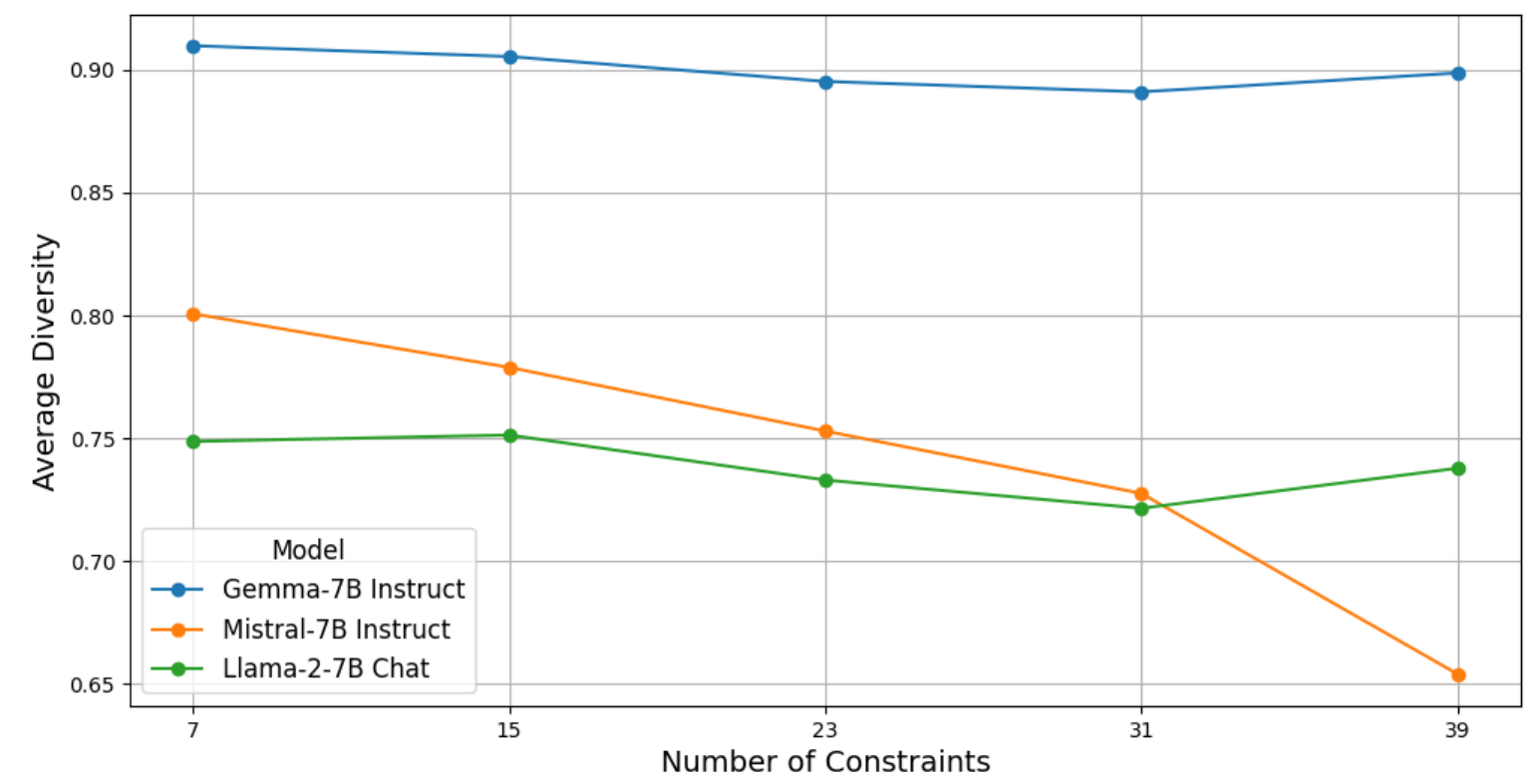}
  \caption{Diversity scores for stories developed using instruction-based constraints.}
  \label{fig:D2Diversity}
\end{figure}

As the prompt specificity increases, two opposing forces could affect the perplexity and diversity of the stories. With more constraints, LLMs have fewer training examples to leverage, resulting in fewer options to generate the next word at any given instance, leading to a low perplexity and diversity score. On the other hand, with more constraints, LLMs are forced to generate more novel stories which would reflect as high perplexity and diversity scores. They could also generate stories with high perplexities by choosing to satisfy only a few of the several constraints in the prompt. To study which of these forces is stronger, it becomes important to compute perplexities and diversity scores for each story. Perplexities are calculated using the same model that generated the story. To analyze the dist-n diversity of the models, we generated three sets of outputs for each input prompt and computed the diversity across these generations. This resulted in a generation of $4500$ stories ($500$ prompts $*$ $3$ LLMs $*$ $3$ stories $=$ $4500$ stories. While Figures~\ref{fig:D3Perplexity}, \ref{fig:OlmoD3Diversity}, and \ref{fig:D3Diversity} show perplexity and diversity results for story-based constraints, Figures~\ref{fig:Type1_Perplexity_combined} and \ref{fig:D2Diversity} show the corresponding results for the instruction-based constraints.

\begin{figure*}[h]
    \centering
    \begin{subfigure}{0.49\textwidth}
        \centering
        \includegraphics[width=\textwidth]{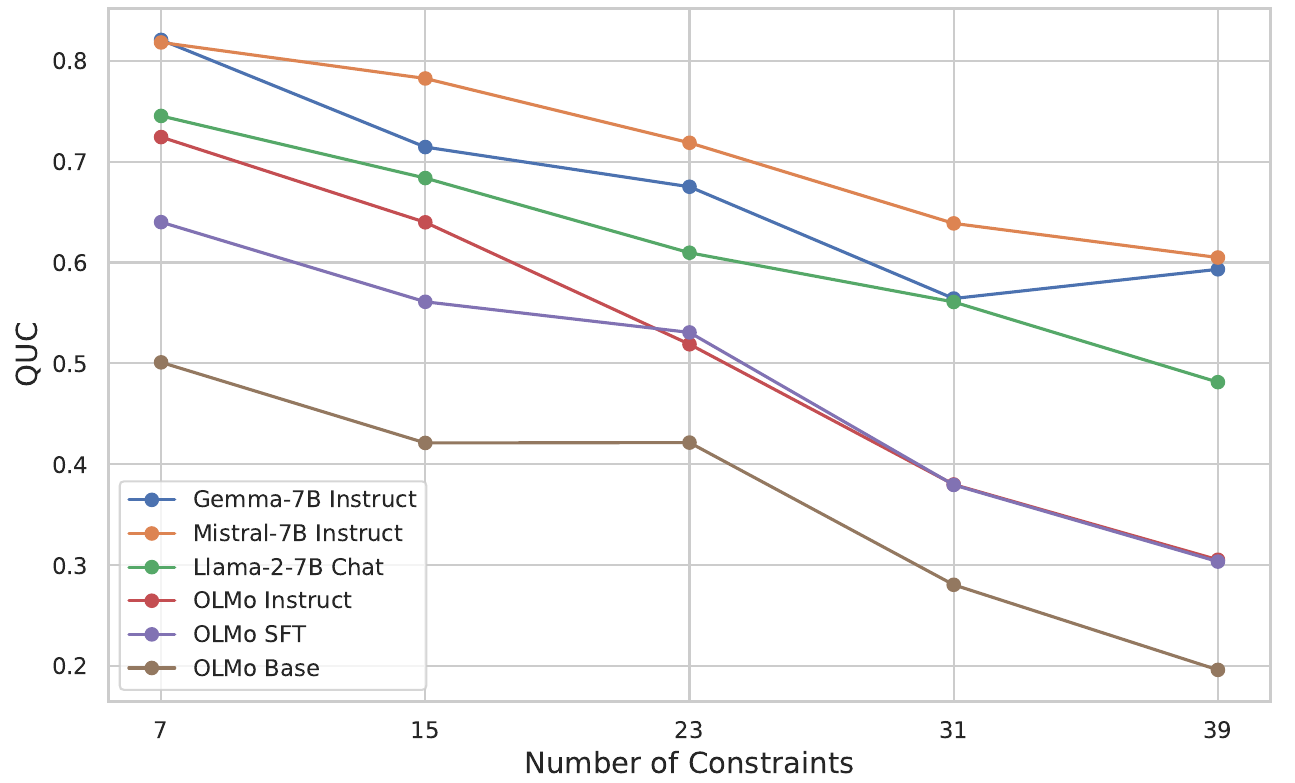}
        \caption{QUC scores for stories developed using instruction-based constraints.}
        \label{fig:All_type1_quc}
    \end{subfigure}%
    \hfill
    \begin{subfigure}{0.49\textwidth}
        \centering
        \includegraphics[width=\textwidth]{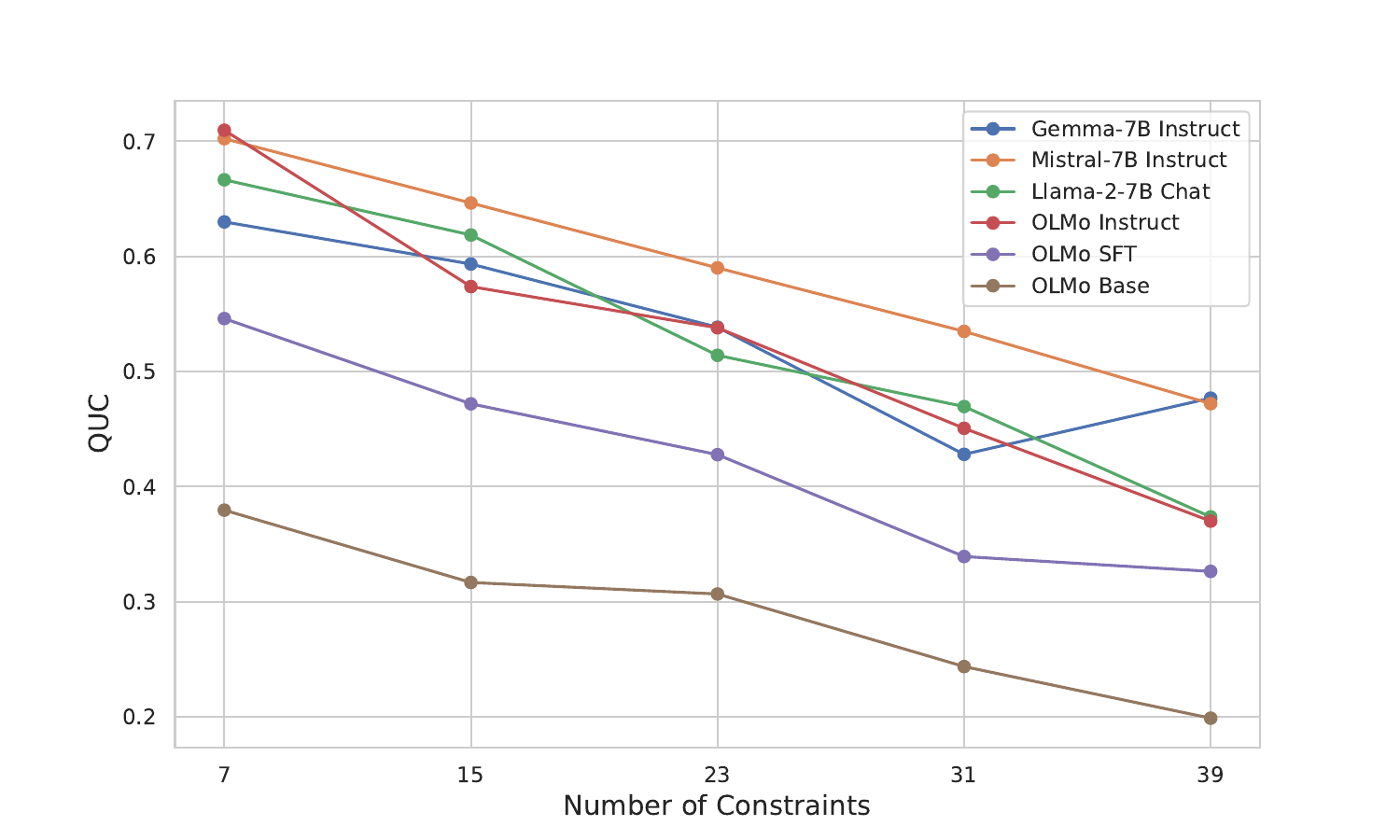}
        \caption{QUC scores for stories developed using story-based constraints.}
        \label{fig:All_type2_quc}
    \end{subfigure}
    
    \caption{QUC scores comparison for stories developed using instruction-based and story-based constraints in \dataset benchmark.}
    \label{fig:QUC_All_Constraints}
\end{figure*}

Finally, \Cref{tab:Type1metrics} denotes the QUC$_{39}$ and RCS$_{7, 39}$ scores for stories generated using $7$ and $39$ instruction-based constraints. These essentially denote the stories generated using the simplest and most complex prompts from \dataset. Here, the story length indicates the number of words in the story which is computed by splitting each story into words using white spaces. In \Cref{tab:Averagedmetrics}, we average the results from \Cref{tab:Type2metrics} and \ref{tab:Type1metrics} to present a more holistic view of how models perform irrespective of the constraint-generation strategy.  The results corresponding to all stories developed from all levels of prompt specificities are shown in \Cref{fig:QUC_All_Constraints}. While OLMo Instruct can be seen to have the steepest decrease in terms of the QUC$_{39}$ score, Mistral and Gemma show relatively consistent performance across different prompt specificities in the \dataset benchmark.

\begin{table}[th!]
\centering
\scalebox{0.77}{
    
    \setlength{\tabcolsep}{2.65pt} % Reduce space between columns
    \begin{tabular}{lcccc}
        \toprule
        & & & \multicolumn{1}{c}{Median} & \multicolumn{1}{c}{Self} \\
        Model       & QUC$_{39}$     &      RCS$_{7, 39}$ $\downarrow$  &            \multicolumn{1}{c}{Length} & \multicolumn{1}{c}{Perplexity}    \\
        \midrule
        Gemma-7B Instruct & 0.5933 & 0.2273 & \textbf{533} & 3.7096\\
        Mistral-7B Chat & \textbf{0.6049} & \textbf{0.2131} & 818 & 1.4958\\
        LLaMA-2-7B Instruct& 0.4814 & 0.2637 & 585 & 2.3285\\
        OLMo-7B Base & 0.1963 & 0.3047 & 374 & \textbf{3.9865}\\
        OLMo-7B SFT & 0.3036 & 0.3363 & 542 & 3.6753\\
        OLMo-7B Instruct & 0.3054 & 0.4188 & 826 & 3.6151\\
        \bottomrule
    \end{tabular}
}
\caption{Overall comparison of different LLMs on instruction-based constraints. Smaller RCS indicates higher creativity. Models adhering closely to the $500$-word limit in our instruction show better compliance and higher self perplexities mean more diverse outputs. Both metrics are computed across all the generated stories. The best values are highlighted.}
\label{tab:Type1metrics}
\end{table}

\begin{table}[th!]
\centering
\scalebox{0.77}
{
    \setlength{\tabcolsep}{2.65pt} % Reduce space between columns
    \begin{tabular}{lcccc}
        \toprule
        & & & \multicolumn{1}{c}{Median} & \multicolumn{1}{c}{Self} \\
        Model       & QUC$_{39}$     &      RCS$_{7, 39}$ $\downarrow$  &            \multicolumn{1}{c}{Length} & \multicolumn{1}{c}{Perplexity}    \\
        \midrule
        Gemma-7B Instruct      & 0.5350 & \textbf{0.1902} & \textbf{510.5}&3.6858 \\
        Mistral-7B Chat       & \textbf{0.5384} & 0.2216& 737.5&1.5396 \\
        LLaMA-2-7B Instruct      & 0.4275 & 0.2783 & 549& 2.3077\\
        OLMo-7B Base    & 0.1975 & 0.2427& 452.5& \textbf{3.8302}\\
        OLMo-7B SFT     & 0.3150 & 0.2779 &557 &3.6462\\
        OLMo-7B Instruct & 0.3377 & 0.3792& 839.5& 3.6079\\
        \bottomrule
    \end{tabular}
}
    \caption{QUC$_{39}$, RCS$_{7, 39}$, story length, and self-perplexity scores averaged across story-based and instruction-based constraints.}
    \label{tab:Averagedmetrics}
\end{table}

%\subsection{Constraint Satisfaction}
%For each constraint, we ask the model to explain if a constraint is not satisfied and to print the sentences of the stories that adhere to the constraint in cases where it is satisfied. 

%\subsection{Story Coherence}
%An example of a single pairwise evaluation is shown in Figure~\ref{fig:Pair-wise}. We don't compare the middle story with itself. The story set as the middle story is compared 14 more times than any other story, and so we divide its coherence score by 14 to provide a fair comparison. 

%\subsection{Output Diversity}

%As the prompt specificity increases, two opposing forces could affect the perplexity and diversity of the stories. With more constraints, LLMs have fewer training examples to leverage, resulting in fewer options to generate the next word at any given instance, leading to a low perplexity and diversity score. 

%On the other hand, with more constraints, LLMs are forced to generate more novel stories which would reflect as high perplexity and diversity scores. They could also generate stories with high perplexities by choosing to satisfy only a few of the several constraints in the prompt. 

%To study which of these forces is stronger, it becomes important to compute perplexities and diversity scores for each story.

\section{Grammar and Likability Scores} 
\label{sec:grammar_likability}
\begin{figure*}[h]
    \centering
    \begin{subfigure}{0.49\textwidth}
        \centering
        \includegraphics[width=\textwidth]{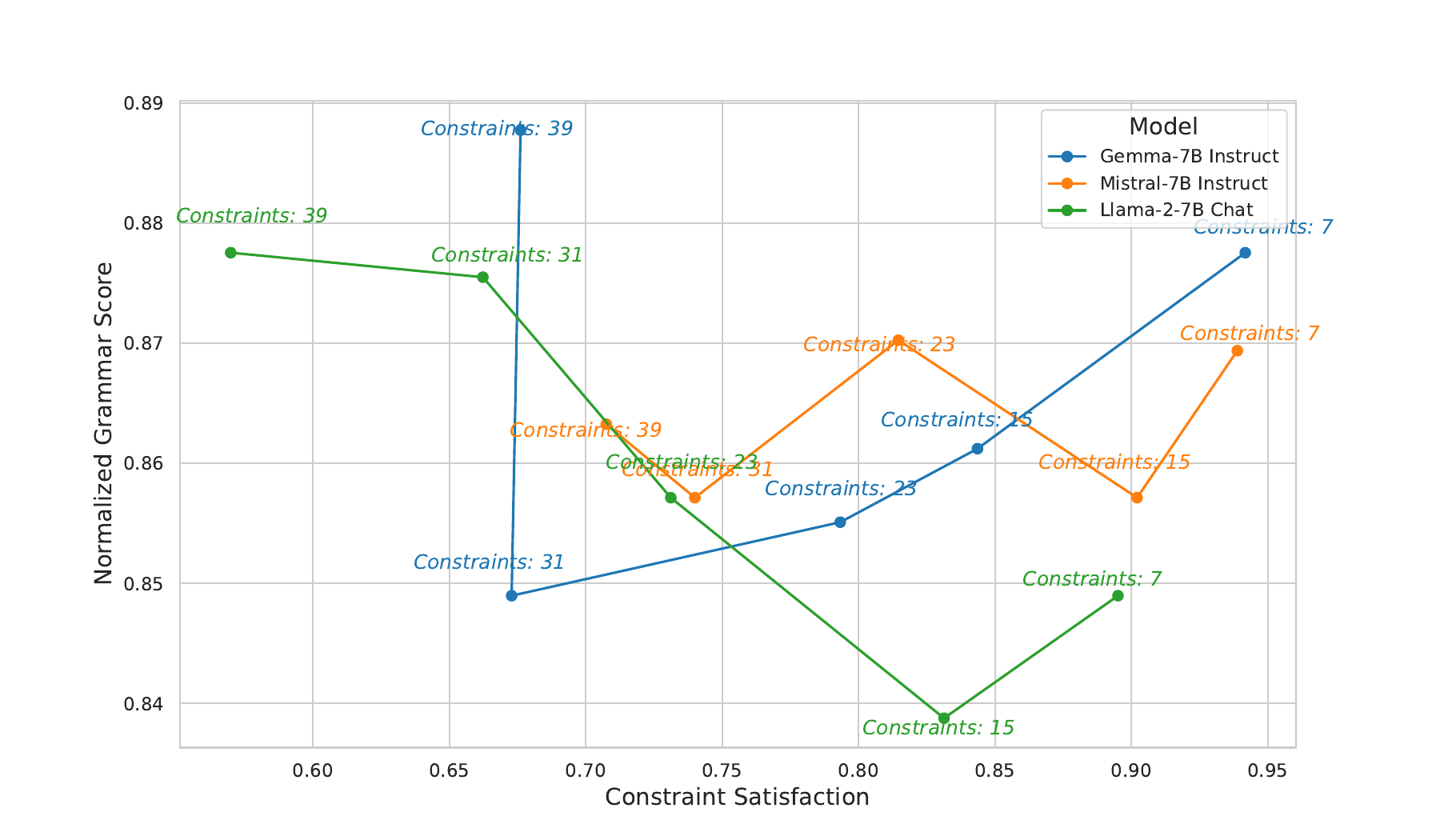}
        \caption{Grammar scores for stories generated using Gemma, Mistral, and LLaMA-2 using instruction-based constraints.}
        \label{fig:grammar_subfig1}
    \end{subfigure}%
    \hfill
    \begin{subfigure}{0.49\textwidth}
        \centering
        \includegraphics[width=\textwidth]{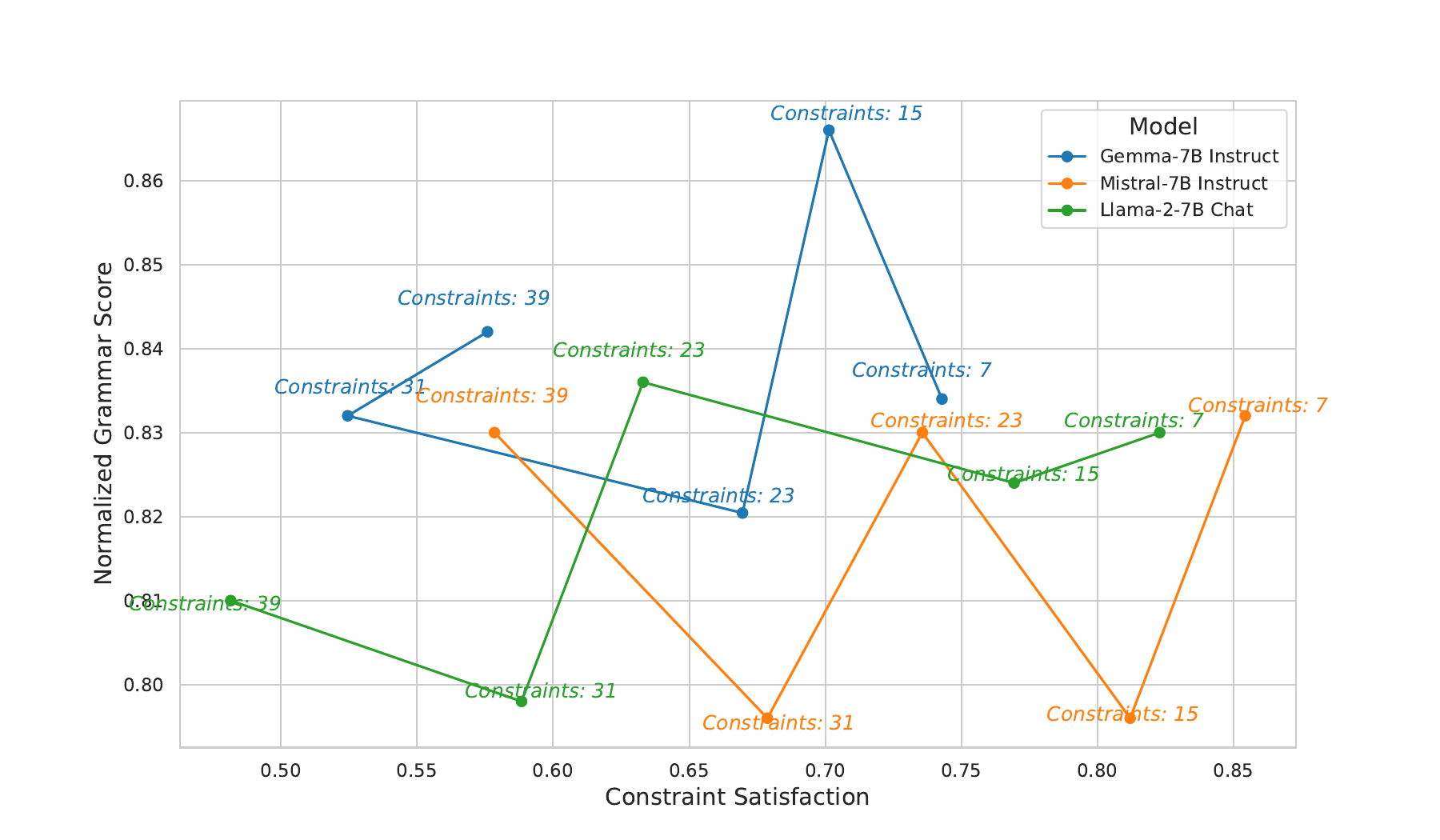}
        \caption{Grammar scores for stories generated Gemma, Mistral, and LLaMA-2 using story-based constraints.}
        \label{fig:grammar_subfig2}
    \end{subfigure}
    
    \vspace{1em} % Adjust vertical space between rows

    \begin{subfigure}{0.49\textwidth}
        \centering
        \includegraphics[width=\textwidth]{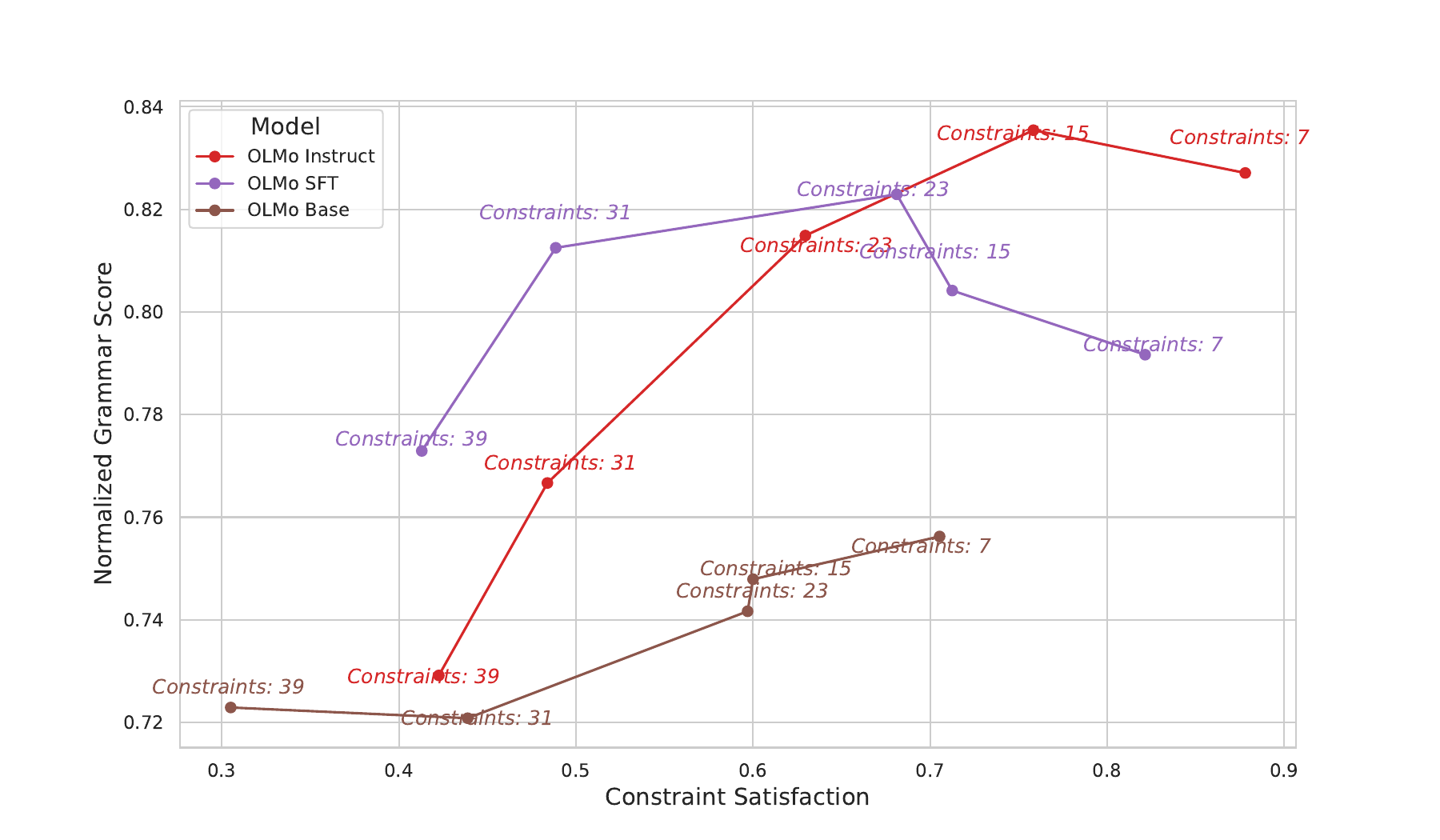}
        \caption{Grammar scores for stories generated using OLMo Base, SFT, and Instruct using instruction-based constraints.}
        \label{fig:grammar_subfig3}
    \end{subfigure}%
    \hfill
    \begin{subfigure}{0.49\textwidth}
        \centering
        \includegraphics[width=\textwidth]{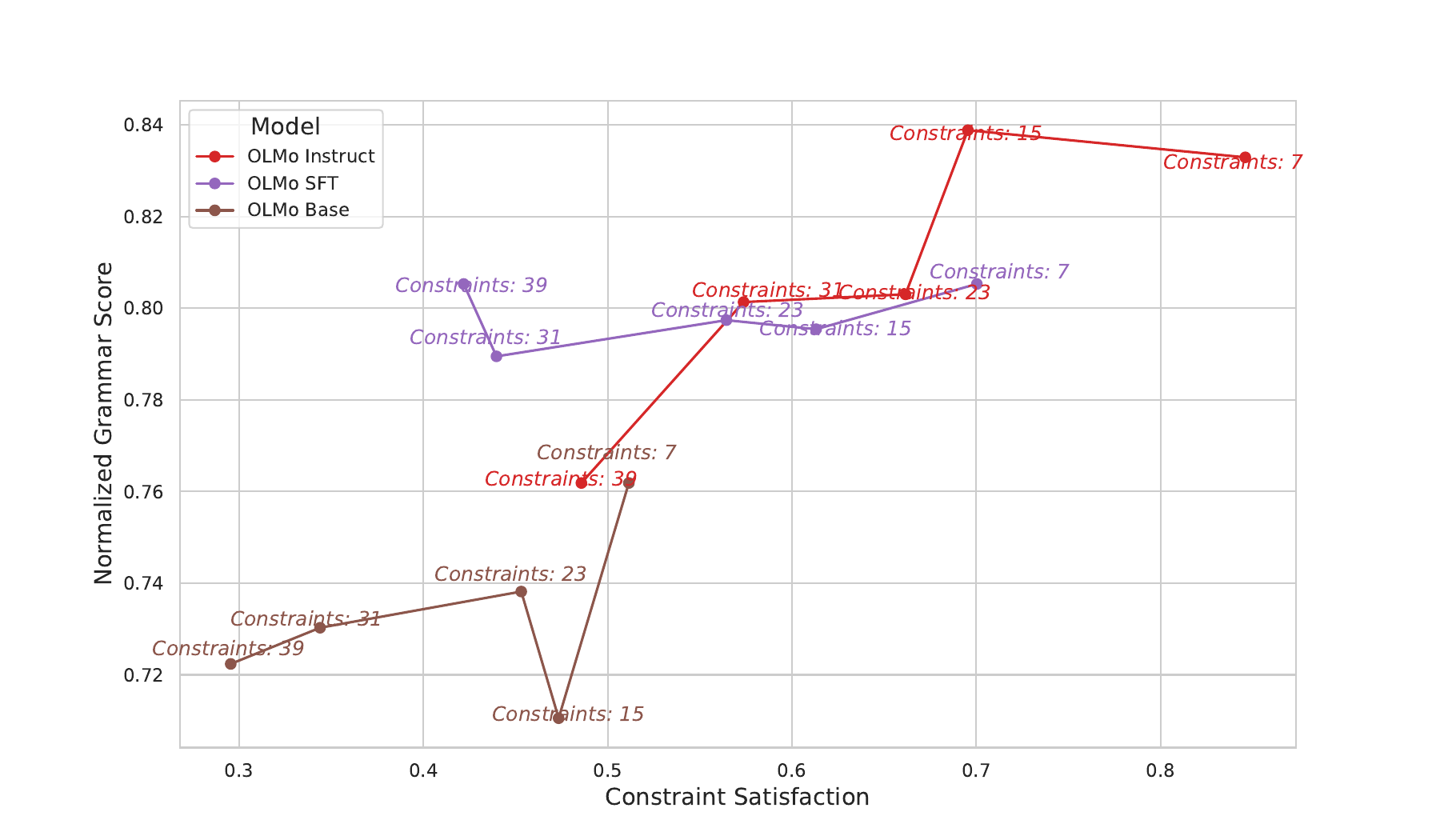}
        \caption{Grammar scores for stories generated using OLMo Base, SFT, and Instruct using story-based constraints.}
        \label{fig:grammar_subfig4}
    \end{subfigure}
    
    \caption{Grammar scores for stories generated using \dataset benchmark.}
    \label{fig:Grammar_combined}
\end{figure*}

\begin{figure*}[h]
    \centering
    \begin{subfigure}{0.49\textwidth}
        \centering
        \includegraphics[width=\textwidth]{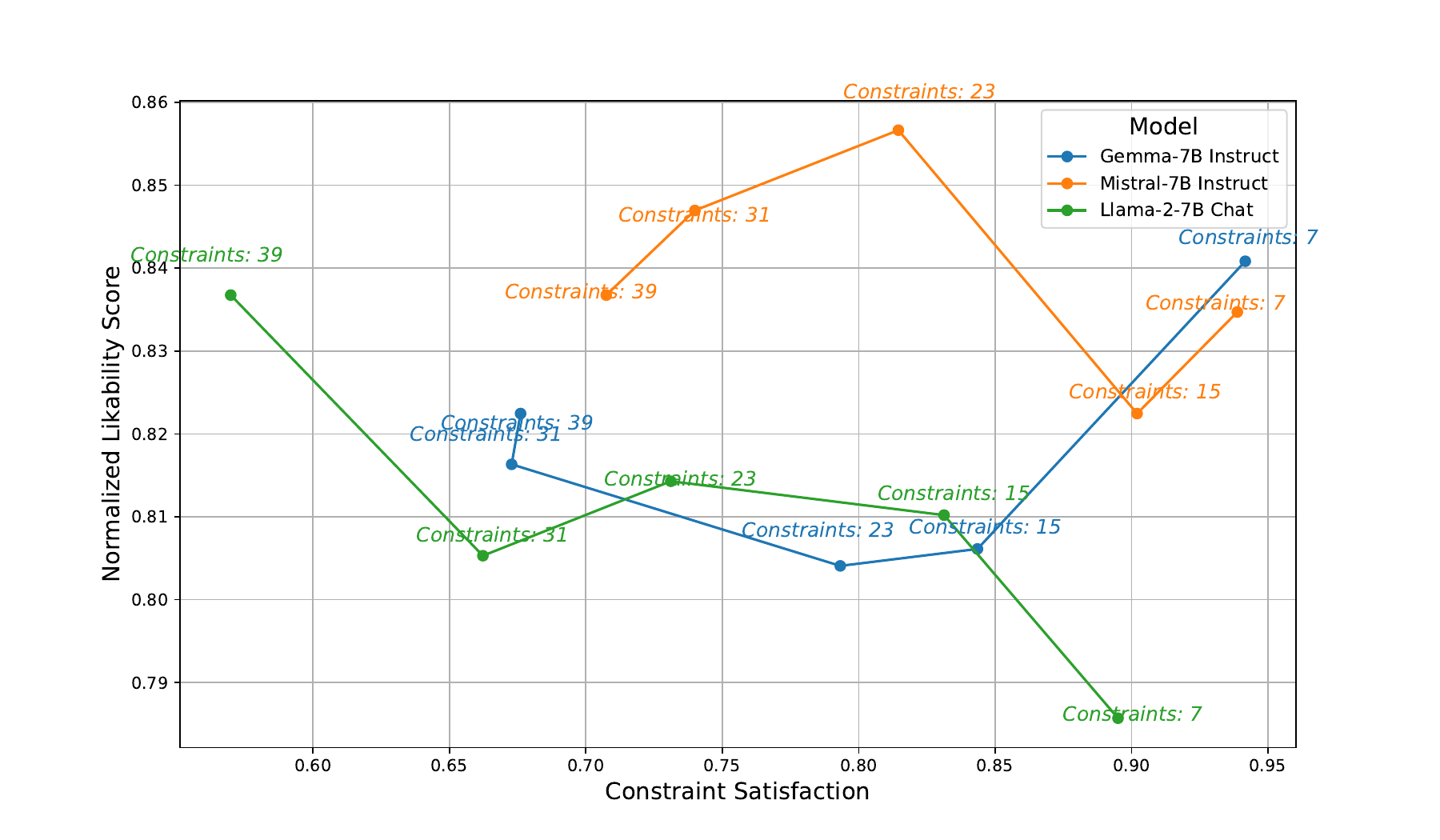}
        \caption{Likability scores for stories generated using Gemma, Mistral, and LLaMA-2 using instruction-based constraints.}
        \label{fig:Likability_subfig1}
    \end{subfigure}%
    \hfill
    \begin{subfigure}{0.49\textwidth}
        \centering
        \includegraphics[width=\textwidth]{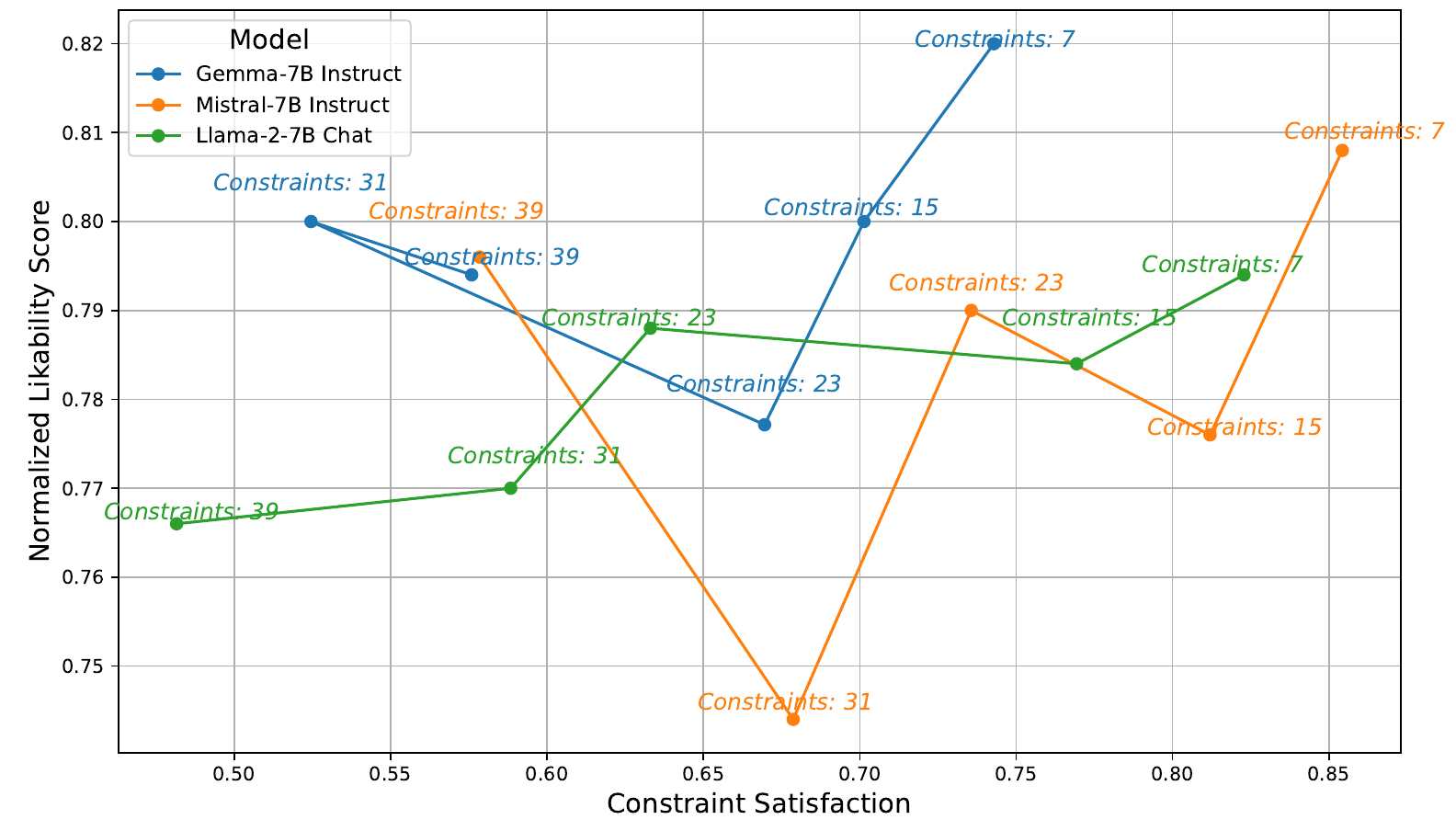}
        \caption{Likability scores for stories generated using Gemma, Mistral, and LLaMA-2 using story-based constraints.}
        \label{fig:Likability_subfig2}
    \end{subfigure}
    
    \vspace{1em} % Adjust vertical space between rows

    \begin{subfigure}{0.49\textwidth}
        \centering
        \includegraphics[width=\textwidth]{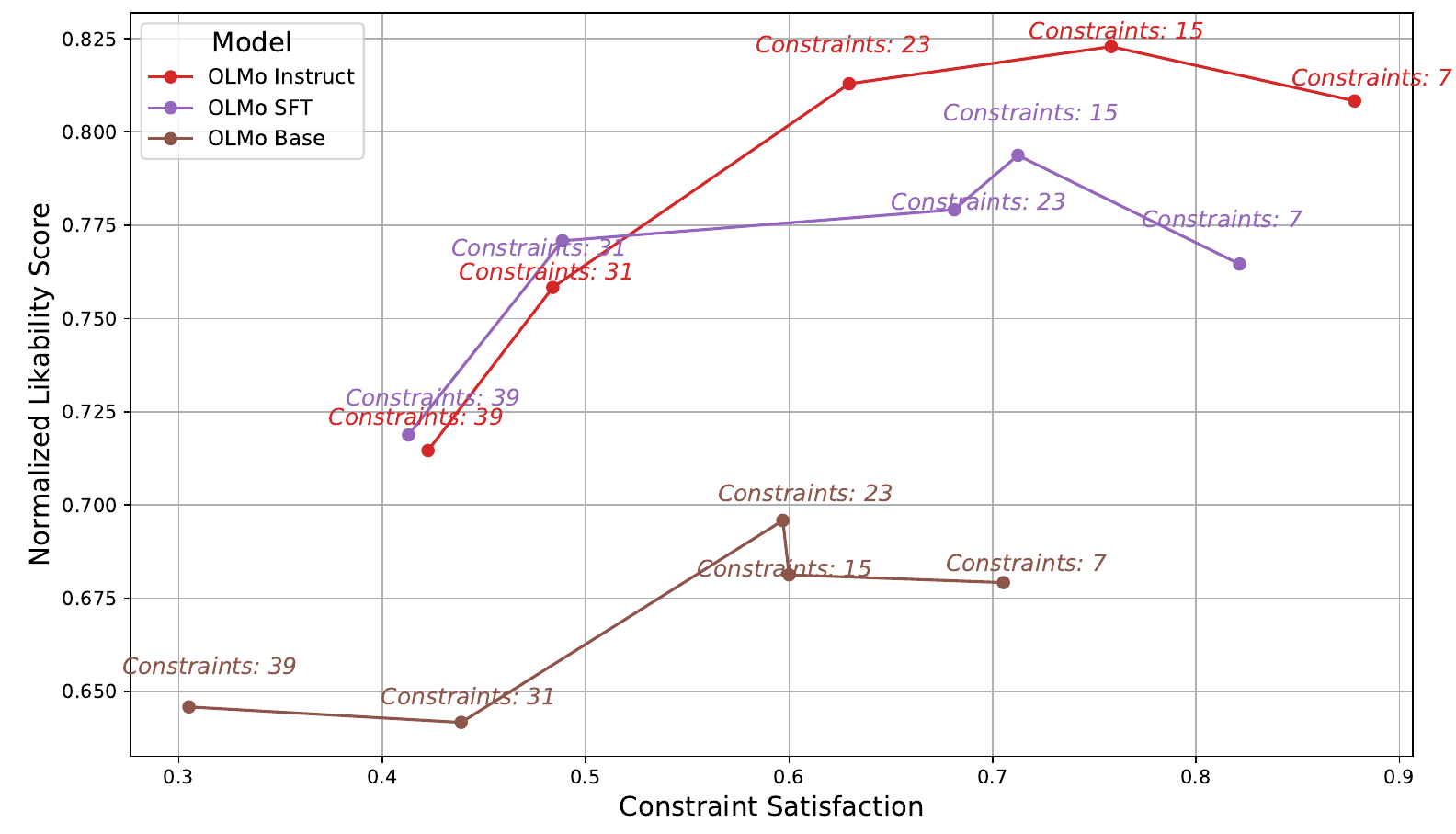}
        \caption{Likability scores for stories generated using OLMo Base, SFT, and Instruct using instruction-based constraints.}
        \label{fig:Likability_subfig3}
    \end{subfigure}%
    \hfill
    \begin{subfigure}{0.49\textwidth}
        \centering
        \includegraphics[width=\textwidth]{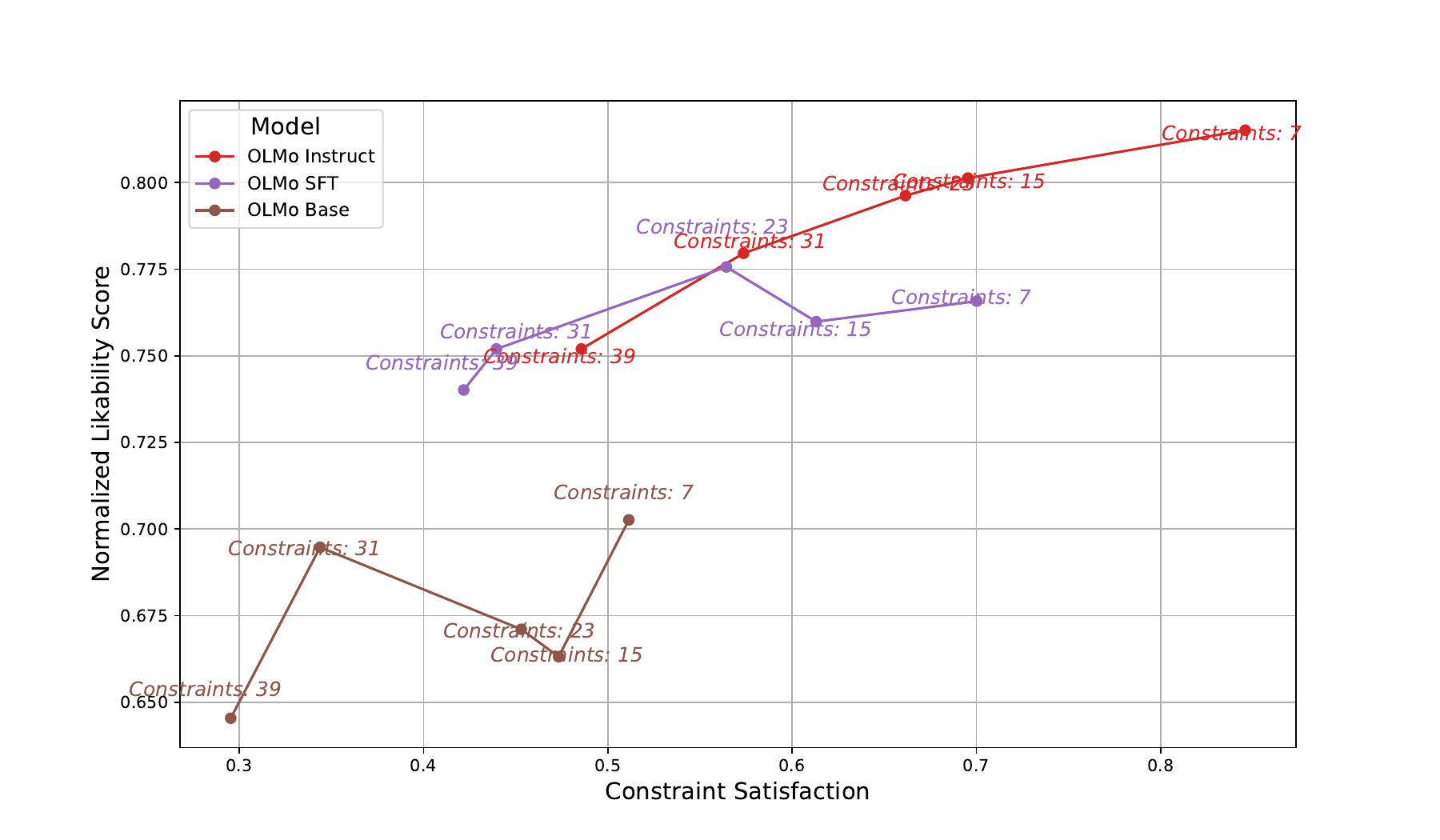}
        \caption{Likability scores for stories generated  OLMo Base, SFT, and Instruct using story-based constraints.}
        \label{fig:Likability_subfig4}
    \end{subfigure}
    
    \caption{Likability scores for stories generated using \dataset benchmark.}
    \label{fig:Likability_combined}
\end{figure*}

Finally, we briefly explain the results obtained while evaluating the stories for their grammar quality and likability. These metrics are computed with coherence using the same pipeline. \Cref{fig:Grammar_combined} shows the normalized grammar scores for stories written using different models with the \dataset benchmark. Similarly, \Cref{fig:Likability_combined} depicts the normalized likability scores. These figures generally indicate that these scores do not vary drastically as the number of constraints increases. While one can see a slight dip in these scores with increased prompt specificity in many cases, these dips are not as significant as the trade-off between constraint satisfaction and coherence. This suggests that models produce stories of similar grammar and likability properties irrespective of the prompt specificity, indicating that these metrics are associated with the model architecture rather than the prompt. For example, one can observe that Mistral, Gemma, and LLaMA-2 always produce competitive grammar and likability scores while OLMo SFT and OLMo Instruct outperform OLMo Base. This means that while LHF can not help produce highly coherent stories in the presence of several constraints, they can still alleviate the model's performance in terms of grammar and likability. 

%____________________

%____________________

\section{Author Contributions}
\begin{itemize}[leftmargin=.1in,topsep=0pt]
\setlength\itemsep{-0.1em}
%\vspace{-0.1em}
    \item \textbf{Anirudh Atmakuru}: A major contributor to constraint generation, LLM evaluation, error analysis, evaluating constraints satisfaction, and paper writing.
    \item \textbf{Jatin Nainani}: A major contributor to defining the project direction, designing evaluation methodologies, performing experiments, and developing the analysis code.
    \item \textbf{Rohith Siddhartha Reddy Bheemreddy}: A major contributor to paper writing, code development, and the creation of story generation and LLM evaluation systems.
    \item \textbf{Anirudh Lakkaraju}: A major contributor to constraint generation, paper writing, and code releasing.
    \item \textbf{Zonghai Yao}: An advisor who contributes to the brainstorming and paper writing.
    \item \textbf{Hamed Zamani}: The faculty advisor who contributes to the paper writing.
    \item \textbf{Haw-Shiuan Chang}: The lead supervisor of the work who proposes the ideas, manages the team, and writes the paper.
\end{itemize}

\end{document}